%% file: main.tex
\documentclass[nohyperref]{article}

\usepackage{microtype}
\usepackage{graphicx}
\usepackage{subcaption}
\usepackage[nointegrals]{wasysym}
\usepackage{booktabs} % for professional tables

\usepackage{hyperref}

\usepackage[accepted]{icml2022}

\usepackage{amsmath}
\usepackage{amssymb}
\usepackage{mathtools}
\usepackage{amsthm}

\usepackage[capitalize,noabbrev]{cleveref}

\theoremstyle{plain}

\theoremstyle{definition}

\theoremstyle{remark}

\usepackage[textsize=tiny]{todonotes}

\usepackage{customnotes}
\icmltitlerunning{Bayesian Imitation Learning for End-to-End Mobile Manipulation}

\begin{document}

\twocolumn[
\icmltitle{Bayesian Imitation Learning for End-to-End Mobile Manipulation}



\icmlsetsymbol{equal}{*}

\begin{icmlauthorlist}
\icmlauthor{Yuqing Du}{berkeley,res}
\icmlauthor{Daniel Ho}{edr}
\icmlauthor{Alexander A. Alemi}{google}
\icmlauthor{Eric Jang}{google}
\icmlauthor{Mohi Khansari}{edr}
\end{icmlauthorlist}

\icmlaffiliation{berkeley}{UC Berkeley}
\icmlaffiliation{google}{Google Research}
\icmlaffiliation{edr}{Everyday Robots}
\icmlaffiliation{res}{Work done while author was at Everyday Robots}

\icmlcorrespondingauthor{Yuqing Du}{yuqing\_du@berkeley.edu}

\icmlkeywords{Machine Learning, ICML}

\vskip 0.3in
]



\printAffiliationsAndNotice{} % otherwise use the standard text.

\begin{abstract}

In this work we investigate and demonstrate benefits of a Bayesian approach to imitation learning from multiple sensor inputs, as applied to the task of opening office doors with a mobile manipulator.
Augmenting policies with additional sensor inputs---such as RGB + depth cameras---is a straightforward approach to improving robot perception capabilities, especially for tasks that may favor different sensors in different situations.
As we scale multi-sensor robotic learning to unstructured real-world settings (e.g. offices, homes) and more complex robot behaviors, we also increase reliance on simulators for cost, efficiency, and safety. Consequently, the sim-to-real gap across multiple sensor modalities also increases, making simulated validation more difficult.
We show that using the Variational Information Bottleneck \cite{alemi2016deep} to regularize convolutional neural networks improves generalization to held-out domains and reduces the sim-to-real gap in a sensor-agnostic manner.
As a side effect, the learned embeddings also provide useful estimates of model uncertainty for each sensor. 
We demonstrate that our method is able to help close the sim-to-real gap and successfully fuse RGB and depth modalities based on understanding of the situational uncertainty of each sensor. In a real-world office environment, we achieve 96\% task success, improving upon the baseline by +16\%.

\end{abstract}

%%%%%%%%% BODY TEXT
\input{introduction}
\input{related_work}
\input{preliminaries}
\input{method}
\input{experiments}
\input{conclusions}

% \bibliography{references}

\input{references.bbl}
\bibliographystyle{icml2022}

%%%%%%%%%%%%%%%%%%%%%%%%%%%%%%%%%%%%%%%%%%%%%%%%%%%%%%%%%%%%%%%%%%%%%%%%%%%%%%%
%%%%%%%%%%%%%%%%%%%%%%%%%%%%%%%%%%%%%%%%%%%%%%%%%%%%%%%%%%%%%%%%%%%%%%%%%%%%%%%
% APPENDIX
%%%%%%%%%%%%%%%%%%%%%%%%%%%%%%%%%%%%%%%%%%%%%%%%%%%%%%%%%%%%%%%%%%%%%%%%%%%%%%%
%%%%%%%%%%%%%%%%%%%%%%%%%%%%%%%%%%%%%%%%%%%%%%%%%%%%%%%%%%%%%%%%%%%%%%%%%%%%%%%
\newpage
\input{appendix}

%%%%%%%%%%%%%%%%%%%%%%%%%%%%%%%%%%%%%%%%%%%%%%%%%%%%%%%%%%%%%%%%%%%%%%%%%%%%%%%
%%%%%%%%%%%%%%%%%%%%%%%%%%%%%%%%%%%%%%%%%%%%%%%%%%%%%%%%%%%%%%%%%%%%%%%%%%%%%%%

\end{document}

%% file: introduction.tex
\section{Introduction}
 A long-standing robotics research problem is to develop agents capable of complex behaviours in the real world. One promising approach is to learn from demonstrations \cite{schaal1997learning}, where the agent learns by modeling the distribution over expert actions. Within the imitation learning paradigm, behavior cloning (BC) \cite{bain1995framework, torabi2018behavioral} is a simple supervised learning method for imitating expert behaviors. In spite of well-known shortcomings, such as compounding errors and the inability to surpass expert performance, recent progress in real-world robotics has shown the promise of BC across different domains and tasks \cite{jang2022bc,  florence2022implicit}.

\begin{figure}[t!]
    \centering
    \includegraphics[width=\linewidth]{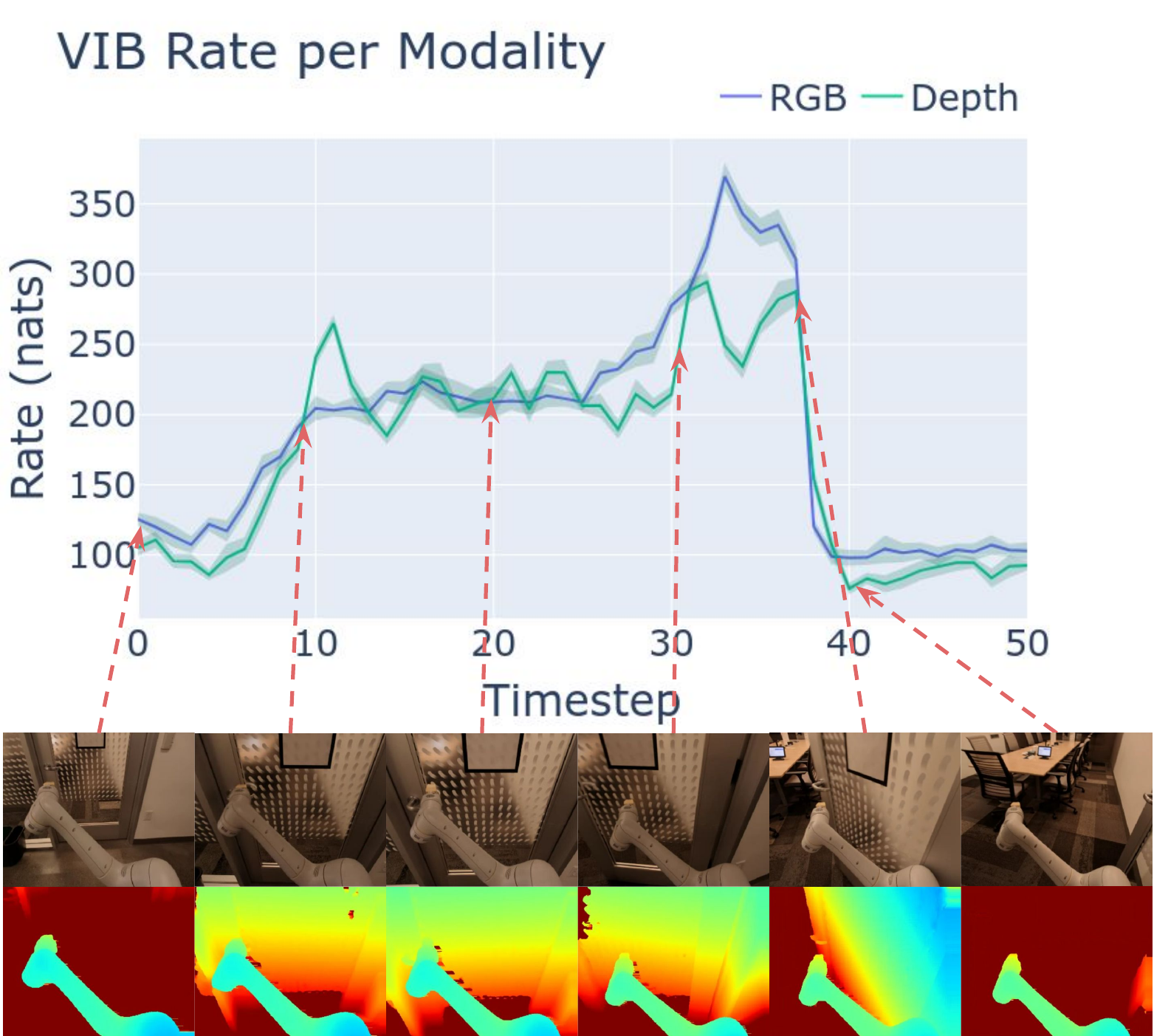}
    \caption{In this work we apply the Variational Information Bottleneck (VIB) to reduce the multi-sensor sim-to-real gap and carry out sensor fusion for a challenging door opening task. The top plot quantifies probabilistic uncertainty of RGB (blue) and Depth (green) sensor modalities as a function of timestep in a sample trajectory, as measured by the VIB rate (divergence of the state representation posterior from its learned marginal). The rate is computed from 8 samples, where the solid line is the mean value and shaded region the standard deviation. The top row of images is RGB observations at each labeled point in time, and the bottom row is the corresponding depth images. In general, we find that the rate is highest when the gripper contacts the door handle or is unlatching the door, for both modalities, suggesting that more information is required for this task-critical phase.}
    \label{fig:trajrate}
\end{figure}
 
 That said, much progress remains to be made towards deploying robots in the real world, especially as we scale to more unstructured, visually diverse environments such as homes or offices. Learning directly in the real world is often costly and challenging. As a result, many end-to-end learning approaches benefit from training and evaluation in simulation. For example, in this work we use simulated evaluations to determine which BC models are suitable for real-world testing, as each real-world evaluation takes significant time, involves numerous scenarios, and requires hands-on human involvement. However, relying on simulation introduces the well known ``reality gap'' \cite{jakobi1995noise}, where policy performance in simulation does not necessarily transfer to the real world. Many approaches have been proposed for tackling this problem \cite{tobin2017domain, sadeghi2016cad2rl}, but they can require significant domain knowledge and engineering. This sim-to-real problem is only exacerbated when multiple sensor modalities are used, as they may each have their own ``reality gap''. Beyond sim-to-real challenges, sensor fusion itself remains an active area of research. Prior work has shown that naive combinations of sensor inputs can hinder policy performance \cite{huang2020multi}. Furthermore, the fusion method itself may suffer from a sim-to-real gap, where a technique that successfully fuses simulated modalities may not extrapolate to reality.

Let us consider the desirable properties of a multi-sensor, end-to-end imitation learning policy.
% that is costly to evaluate in the real world. 
The policy should: 1) be invariant to sim and real domains on a per-sensor level, 2) quantify when each sensor representation is "uncertain" and rely on it less, and 3) be generally applied to combinations of different sensor modalities. We propose that introducing a information bottleneck, specifically, the Variational Information Bottleneck \cite{alemi2016deep}, meets these requirements: 1) a bottleneck with finite channel capacity may force sim and real domains to be encoded using shared bits, 2) VIB has been shown to yield calibrated uncertainty estimates to out-of-distribution examples \cite{alemi2018uncertainty}, and 3) VIB can be dropped in as an additional layer and loss without modifying the rest of the architecture. To the best of our knowledge, VIB has not been demonstrated to yield calibrated OoD detection capabilities across multiple modalities on real-world robotic systems to date.

We choose a challenging mobile manipulation task in the real world---latched door opening in an unstructured office environment---as our testbed. Door opening is a required capability for any general-purpose mobile robot performing tasks in human environments like homes and offices. Our policy should be able to bridge the sim-to-real gap broadened by constantly changing real office spaces, successfully make use of all available sensor modalities, and handle the combined complexities of manipulation and navigation.

We investigate our hypotheses about the applicability of VIB to improving generalization in large-scale robotic imitation learning as well as sensor fusion. We train a separate stochastic encoder for each sensory input on both sim and real domains, hypothesizing that the VIB objective encourages bits encoding sim and real features to be shared in a domain-agnostic manner, while still being predictive of demonstrated actions. As an ancillary benefit, this approach also admits tractable probabilistic representations of model uncertainty. We find that the learned VIB rates (the KL divergence of the state-posterior from the state-prior, in nats) are useful for estimating which sensor is more reliable, and thus they can be used in a simple softmax-weighted sensor fusion scheme. Our method uses the actions predicted by the modality whose input has the lowest `model uncertainty'. 
 
Our contributions are as follows: 
\begin{itemize}
     \item We define a behavior cloning approach that uses VIB for learning domain-agnostic embeddings such that we are able to close the sim-to-real gap.
     \item We demonstrate that our learned embeddings form a meaningful latent space such that the per-instance VIB rate is informative of task-significant inputs, and that the VIB rate can be used as a measure of modality-specific uncertainty for explainable sensor fusion.
     \item We tackle the challenging robotics problem of latched door opening in unstructured real-world environments, achieving 96\% success.
\end{itemize}

%% file: related_work.tex
\section{Related Work}

\textbf{Information Bottleneck for Control.} The information bottleneck (IB) \cite{tishby2000information} was originally proposed as a method for finding a compressed representation of the input signal that preserves maximum information about the desired output signal. \citet{alemi2016deep} extend the IB approach to deep networks by using a variational lower bound of the IB objective and using the reparametrization trick \cite{kingma2013auto}. While originally studied in the context of image classification, representation learning using the VIB objective has also been applied to other domains such as unsupervised learning (e.g. $\beta$-VAE \cite{higgins2016beta}), meta-learning \cite{du2020learning}, and control.

In the domain of control, prior work has explored using the IB principle to learn compressed representations. In reinforcement learning, this includes learning task-critical representations \cite{pacelli2020learning, lu2020dynamics}, improving the stability of actor-critic methods \cite{igl2019generalization}, tackling the exploration problem \cite{goyal2019infobot}, and addressing overfitting in offline RL \cite{kumar2021workflow}. Closer to our work in the context of imitation learning, \citet{peng2018variational} propose the Variational Discriminator Bottleneck (VDB) for regularizing the discriminator in adversarial learning methods. While they apply the VDB for adversarial imitation learning in simulation \cite{ho2016generative}, our work instead focuses on behavior cloning for real world robotics and avoids the complexities of adversarial training. \citet{lynch2021language} propose LangLFP, a multicontext conditional VAE-based imitation learning policy that learns a representation invariant to both visual and text modalities. However, they do not explicitly use the uncertainty estimates for modality fusion. To the best of our knowledge, our work is the first visual end-to-end imitation learning work to demonstrate the efficacy of the VIB for both sim-to-real transfer and multi-sensor fusion. 

\textbf{Sim-to-Real.} Sim-to-real transfer allows a model trained in a simulated domain to perform well in the real world. This can be accomplished by reducing the reality gap by making a simulation environment more similar to the real world, or, by learning a model representation that is domain-agnostic or robust across many domains.

In this work, we focus on the principle of domain adaptation---in which input from disparate domains are adapted to be more similar--- specifically by inducing a domain-agnostic feature representation of the input. The works DANN and DSN \cite{DANN, dsn} adversarially teach a network to extract features which do not discriminate between sim and real domains. Feature-level adaptation can be conceptually similar to other self-supervised representation learning work, which also aims to increase similarity between embeddings of positive image pairs. These positive pairs have been generated from image augmentations, patches, and color \cite{chen2020simple, henaff2020dataefficient, chen2020exploring,pathak2016context, mundhenk2018improvements, noroozi2016unsupervised, zhang2017split}---concepts which perturb the input but not ground truth labels, or leverage other invariants based on the input state. In this work, we instead rely on an information bottleneck to learn a shared, compressed representation for both simulated and real domains, without requiring image pairs. 

\textbf{Multi-Sensor Fusion.} Many prior works have shown that increasing sensor modalities (e.g. tactile, audio, visual) can improve control for domains such as manipulation \cite{lee2020making} and autonomous navigation \cite{fayyad2020deep}. However, how to best combine multiple modalities is an active area of research.
 
 Successful fusion at the representation level for end-to-end imitation learning remains challenging due to causal confusion \cite{codevilla2019exploring}. The network can learn to ignore or overly focus on a modality due to spurious correlations that can occur when demonstrations are the only supervision signal \cite{huang2020multi}. Causal confusion is not limited to the multi-sensor regime; simply increasing the dimensionality of the inputs can also increase the number of spurious correlations. One way to overcome this challenge is to use auxiliary losses to enforce a more meaningful multimodal representation \cite{xu2017end, lee2020making}; however, developing such auxiliary losses can require task and sensor specific loss engineering. 
 
 More similar to our work is a rich literature of approaches that aim to account for sensor uncertainty during fusion. These include probabilistic methods \cite{proencca2018probabilistic, murphy1998dempster}, modeling sensor noise distributions \cite{zhu2013variational}, 
 and learned confidence maps \cite{van2019sparse}. However, these prior approaches are not applied to end-to-end learning. Furthermore, we introduce the insight of using an inherent measure of uncertainty from the VIB for explainable sensor fusion.
 

%% file: preliminaries.tex
\section{Preliminaries}

Given an input source $X$, stochastic encoding $Z$, and target variable $Y$, the Information Bottleneck (IB) \cite{tishby2000information} approach optimizes for an encoding $Z$ that is maximally predictive of $Y$ while being a compressed representation of $X$. A parametric encoder $p(z|x; \theta)$, typically modeled via a neural network with weights $\theta$, is optimized through
\begin{equation}
    \max_\theta \;\; I(Z; Y | \theta) - \beta I(Z;X | \theta)
\end{equation}
where $I(A;B)$ is the mutual information between variables $A,B$, formally defined as the KL-divergence between the joint density and the product of the marginals,
\begin{equation}
    I(A;B) = D_{KL} [p(a,b) || p(a) p(b)]
\end{equation}
The hyperparameter $\beta$ controls the tradeoff between the predictive power and degree of compression of $z$, where $\beta=0$ corresponds to a stochastic version of the typical maximum likelihood objective. However, as mutual information is generally computationally intractable, \citet{alemi2016deep} propose the Variational Information Bottleneck (VIB) to learn a variational lower bound of the IB, extending the IB method to deep neural networks and high dimensional inputs. The VIB objective to be maximized takes the form:
\begin{equation}\label{eq:VIB}
     \frac{1}{N}\sum_{n=1}^N  \mathbb{E}_{z \sim p(z|x_n)} \left[\log q(y_n|z)  - \beta \log \frac{p(z|x_n)}{r(z)} \right] 
\end{equation}
 where $p(z|x)$ is our encoder, $q(y|z)$ is a variational approximation to $p(y|z) = \int dx\, p(y|x)p(z|x)p(x)/p(z)$, $r(z)$ is a variational approximation of $p(z) = \int dx\, p(z|x)p(x)$, and $N$ is the number of training examples. As is done in standard practice, we parametrize the encoder distribution $p(z|x)$ as a multivariate Gaussian, where the mean and diagonal terms of the covariance matrix are predicted by a neural network. We use the reparametrization trick to compute derivatives of the network parameters with respect to losses on the stochastic samples. As computing the marginal distribution $p(z)$ exactly is intractable, $r(z)$ is often parametrized by a learned model as well. In our case, we model $r(z)$ as a mixture-of-Gaussians with the number of modes as a hyperparameter. 

%% file: method.tex
\begin{figure*}[ht]
    \centering
    \includegraphics[width=.83\linewidth]{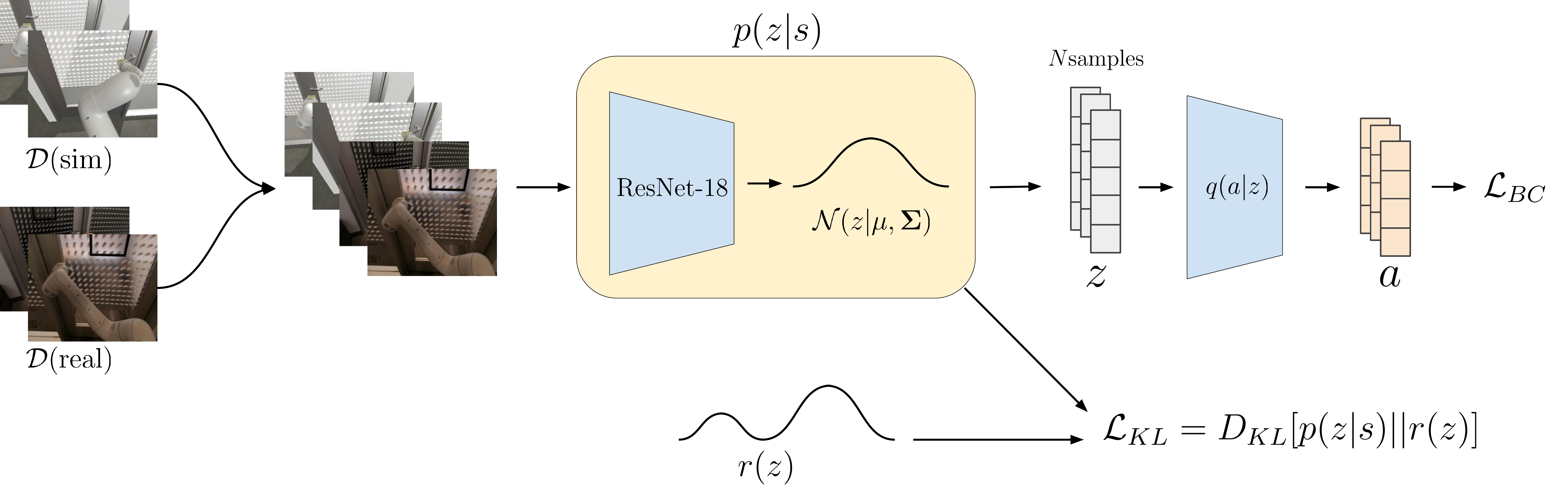}
    \vspace{-5mm}
    \caption{Overall model diagram for a single modality. We learn a stochastic encoder $p(z|s)$, an action decoder $q(a|z)$, and a prior $r(z)$. The objective (Eq. \ref{eq:bc_vib}) consists of both a behaviour cloning loss and a KL-divergence term for imposing the Variational Information Bottleneck (VIB). We train all components using a mixed sim and real dataset to encourage the latent representation to be domain agnostic.}
    \label{fig:modeldiag}
\end{figure*}

\section{Bayesian Imitation Learning}

In imitation learning, we are provided with a set of expert demonstrations, consisting of input observations $s$ and corresponding expert actions $a$. Our goal is to learn a policy $\pi(a|s)$ that replicates the expert. To make use of the VIB, we decompose the policy into a stochastic encoder, $p(z|s)$, and an action decoder, $q(a|z)$, and we impose a bottleneck on the learned stochastic encoding $z$ using Eq. \ref{eq:VIB}. For a single input image $s$, we can decompose the training loss as 

\vspace{-5mm}

\begin{equation}\label{eq:bc_vib}
    \mathcal{L} = \underbrace{\mathbb{E}_{z \sim p(z|s)} \left[-\log q(a|z)\right]}_{\mathcal{L}_{BC}}  + \beta \underbrace{\mathbb{E}_{z \sim p(z|s)} \left[\log \frac{p(z|s)}{r(z)} \right]}_{\mathcal{L}_{KL}} 
\end{equation}
\vspace{-6mm}

where the first term $\mathcal{L}_{BC}$ is the behaviour cloning loss. Following \citet{jang2022bc}, we use a Huber loss \cite{huber} between the predicted and demonstrated actions instead of a negative log-likelihood loss on an explicit distribution. The second term is the rate $\mathcal{L}_{KL}$, equivalent to $D_{KL}[p(z|x)||r(z)] \geq I(X;Z)$. The latter is weighted by $\beta$ and controls the bottlenecking tradeoff. We use simulated evaluations for the $\beta$ hyperparameter sweep, as real evaluations are costly and time-consuming. As observed in \citet{lu2020dynamics}, in practice we find it helpful to linearly anneal $\beta$ from 0 within the first 3000 steps of training.

In training our policy with Eq. \ref{eq:bc_vib}, we aim to learn an encoding space $z$ that is maximally predictive of the expert actions while being maximally concise about the input observations. Since the encoder is trained on both simulated and real images, we hypothesize that the shared yet compressed representations will help close the sim-to-real gap.

\textbf{Network Details.}  We parameterize the stochastic encoder $p(z|s)$ using a ResNet-18 \cite{he2015deep} that predicts the mean and covariance of a multivariate Gaussian distribution on $\mathbb{R}^{64}$. 
The action decoder $q(a|z)$ is a 2-layer MLP, and the learned prior $r(z)$ is a mixture of multivariate Gaussians on $\mathbb{R}^{64}$ with 512 components and a learnable mean and diagonal covariance matrix. 

We estimate both expectations in Eq. \ref{eq:bc_vib} using Monte Carlo sampling. This is necessary for the second term as we parametrize our prior with a multivariate Gaussian mixture model, from which the KL divergence is analytically intractable \cite{hershey2007approximating}. During training we take 8 samples from the stochastic embedding per input and compute the average VIB rate loss. Similarly, we decode each sample into separate actions and compute the average behavior cloning loss. At inference time, we execute the mean action prediction for a modality by computing the model average across the samples.

\subsection{Uncertainty-based Sensor Fusion}\label{sec:fusion}
Sensor fusion is a well-known challenge in robotics. While increasing the number of sensors increases the amount of information a system can glean from the environment, composing multiple sensor inputs can be problematic, especially if they are in disagreement. Ideally, we would like our multi-sensor system to rely less on inputs it is uncertain about (e.g. dissimilar to the expert data). For example, if the RGB camera records significantly different colors, lights, or objects than those in the training dataset, the model should rely on action predictions from the depth modality instead.

As an estimate of sensor uncertainty, we make use of the  per-instance rate $\mathcal{L}_{KL}$. Similarly to \citet{alemi2018uncertainty}, we expect higher rates to correlate with higher uncertainty about the input, which can also correspond to inputs uncommon in the training dataset (eg. out-of-distribution inputs).
In the context of sensor fusion, we execute the actions from the modalities with lower rates (i.e. more `in-distribution'), allowing us to fuse sensors in an uncertainty-aware manner.

\begin{figure}[t!]
    \centering
    \includegraphics[width=.88\linewidth]{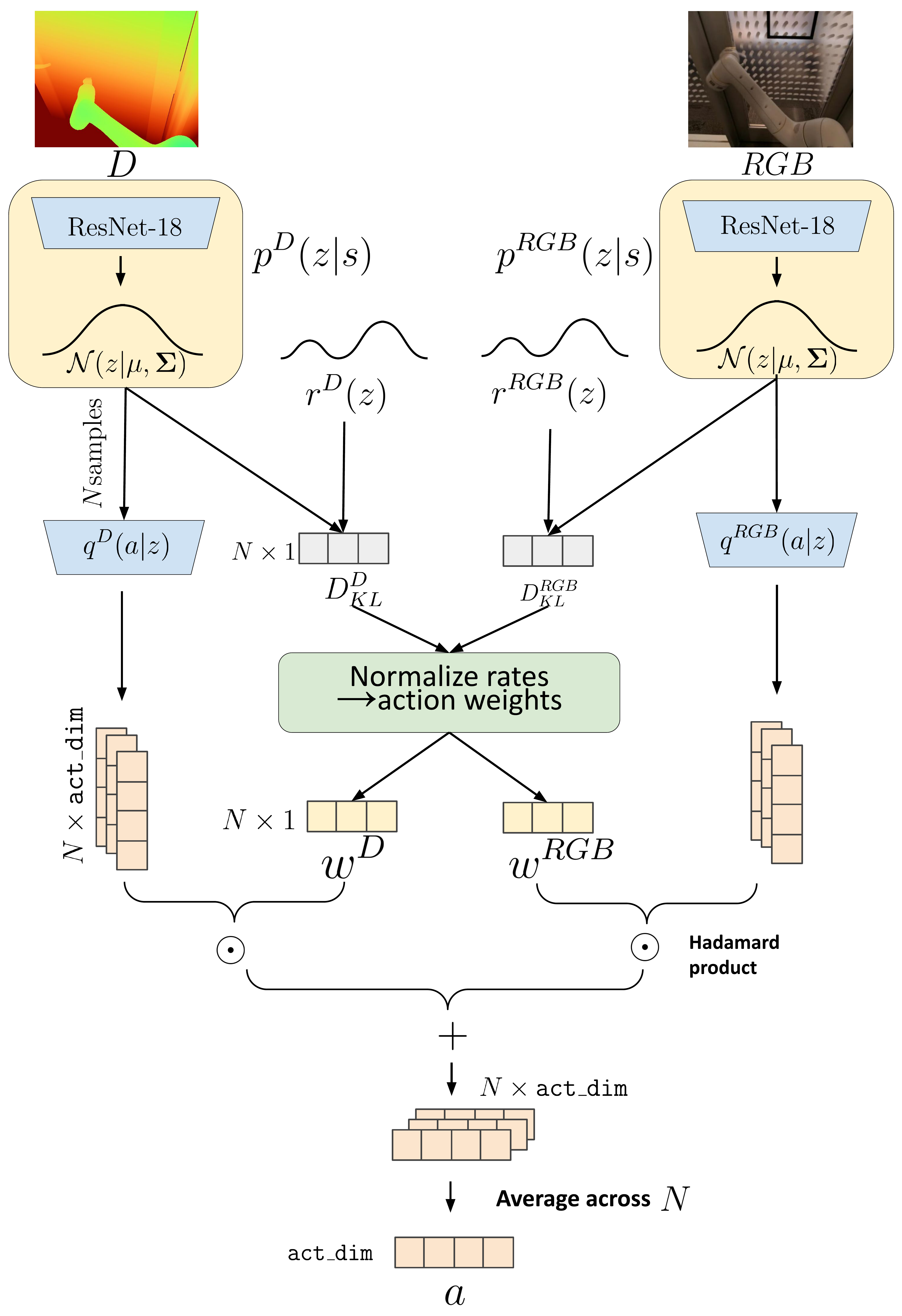}
    \vspace{-5mm}
    \caption{System diagram showing how each modality contributes to the final action prediction. Each sensor is trained to have an independent encoder $p(z|s)$, decoder $q(a|z)$, and prior $r(z)$. During inference, we weigh the contribution of each modality's action prediction using the VIB rate such that modalities with lower rates (correlated with lower uncertainty) are in more control.}
    \vspace{-2mm}
    \label{fig:fusion}
\end{figure}

Given $N$ sensor inputs, we train $i = 1, ..., N$ models independently. That is, each modality's model has its own encoder $p_i(z|x)$, decoder $q_i(a|z)$, and learned marginal $r_i(z)$. This allows us to impose modality-specific bottlenecks, as the ideal lower-dimensional density model may vary across modalities. Each model only sees the input image corresponding to its designated modality. At inference time, we use the VIB rate, $\mathcal{L}_{KL}^i$, of the $i$th model as a measure of modality uncertainty.  Since we want lower rates to correspond to higher action weights, we first compute the unnormalized weights for the $j$th model as $\bar{w}^j = \sum_i \mathcal{L}_{KL}^i - \mathcal{L}_{KL}^j$. We then normalize the weights per modality $w^i$ such that $\sum_i w^i = 1$. We experiment with two normalization schemes: 1) softmax normalization for more discrete modality switching, and 2) dividing  $\bar{w}^i$ by $\sum_i \mathcal{L}_{KL}^i$ for more blended actions. We compute the fused action as $a = \sum_i w^i a^i$. While we evaluate the two most common image modalities for robotics, RGB and depth, in principle our proposed method can be extended to additional modalities. Figure \ref{fig:fusion} gives an overview of the fusion process.

%% file: experiments.tex
\section{Experiments}
Our experiments aim to answer the following questions: 1) Does the regularization imposed by the information bottleneck lead to domain-agnostic representations, thus closing the sim-to-real gap? 2) Can the VIB-based representation help with making the model more explainable? 3) Can the VIB rates across modalities be used for uncertainty-based sensor fusion? 

\subsection{Experimental Setup}
Our training dataset consists of a real-world dataset of 2068 demonstrations ($\sim$13.5 hours) and a simulated dataset of $\sim$500 demonstrations ($\sim$2.7 hours), all collected using hand-held teleoperation devices. The real-world dataset does not control for the interior of the room, leaving the observations as natural as possible. We also train a RetinaGAN model \cite{retinagan} on the sim and real dataset, and use the GAN model to translate sim images to look like real and vice-versa. We use all four datasets (sim, adapted sim, real, and adapted real) to impose a feature-level consistency loss \cite{khansari2022practical} across all models (including baselines).

We carry out evaluations on 10 different latched doors, split into 5 left swinging and 5 right swinging. Six of the doors are in the training dataset and four are only used during evaluation---entirely unseen during training. Each door is tested with 30 trials split across two different robots, of which only one was used to collect the training dataset. As in the training dataset, we do not control for the interior of the rooms (eg. objects in the room, pose of furniture).

For each model configuration, we train three separate models that differ only by the random training seed. From these, we choose three checkpoints with the highest evaluation performance in simulation to use for real evaluation. During real evaluation, we randomly sample from the three checkpoints in a manner blind to the robot operator. We report estimated standard deviation as $\sqrt{p(1-p)/(n-1)}$, assuming the $n$ trials are i.i.d. Bernoulli variables with success $p$. For details on the training setup and evaluation protocols, see \cite{khansari2022practical}.

\subsection{Results and Discussion}

\textbf{Q1. Domain-Agnostic Representations.}
 Since we use the same encoder for both simulated and real images of the same modality, our goal is to learn a bottlenecked encoding that is domain agnostic, i.e. helps close the sim-to-real gap. Since our approach enables us to learn a lower-dimensional density model of the input images, we can directly investigate whether this is the case. To do so, we measure the KL divergence between the distributions parametrized by the embeddings of simulated and real images. 

\begin{figure*}[h!]
\begin{subfigure}{\linewidth}
  \centering
  % include first image
  \includegraphics[width=\linewidth]{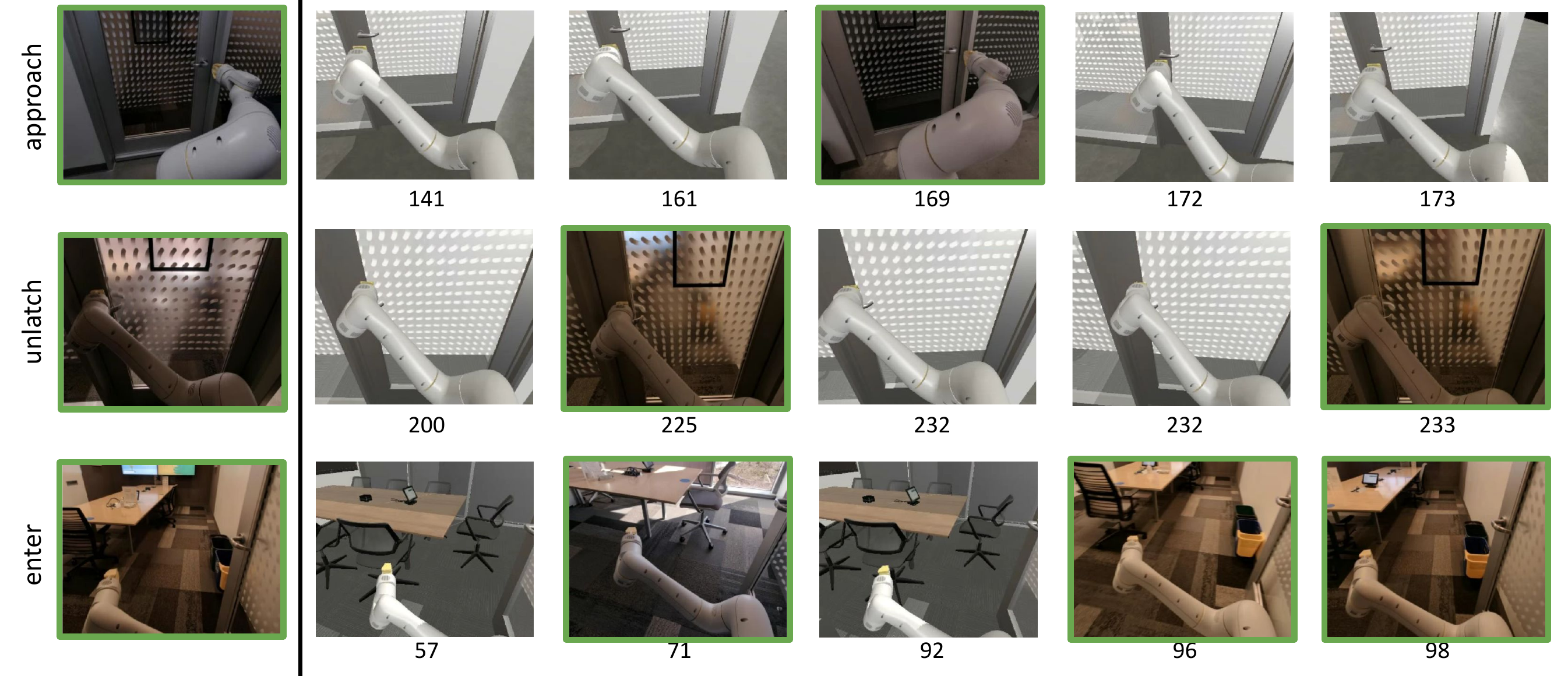}  
  \caption{Real-world anchor images. First row: first phase of door opening moving towards door; second row: manipulating the door handle to unlatch; third row: navigating into the room. The closest images in the first row are all during the approaching phase, varying in arm orientation and includes both sim/real domains. The second row shows the same door unlatching configuration, both sim/real domains. The last row shows different room interiors, both sim/real domains.}
  \label{fig:sub-first}
\end{subfigure}
\begin{subfigure}{\linewidth}
  \centering
  % include second image
  \includegraphics[width=\linewidth]{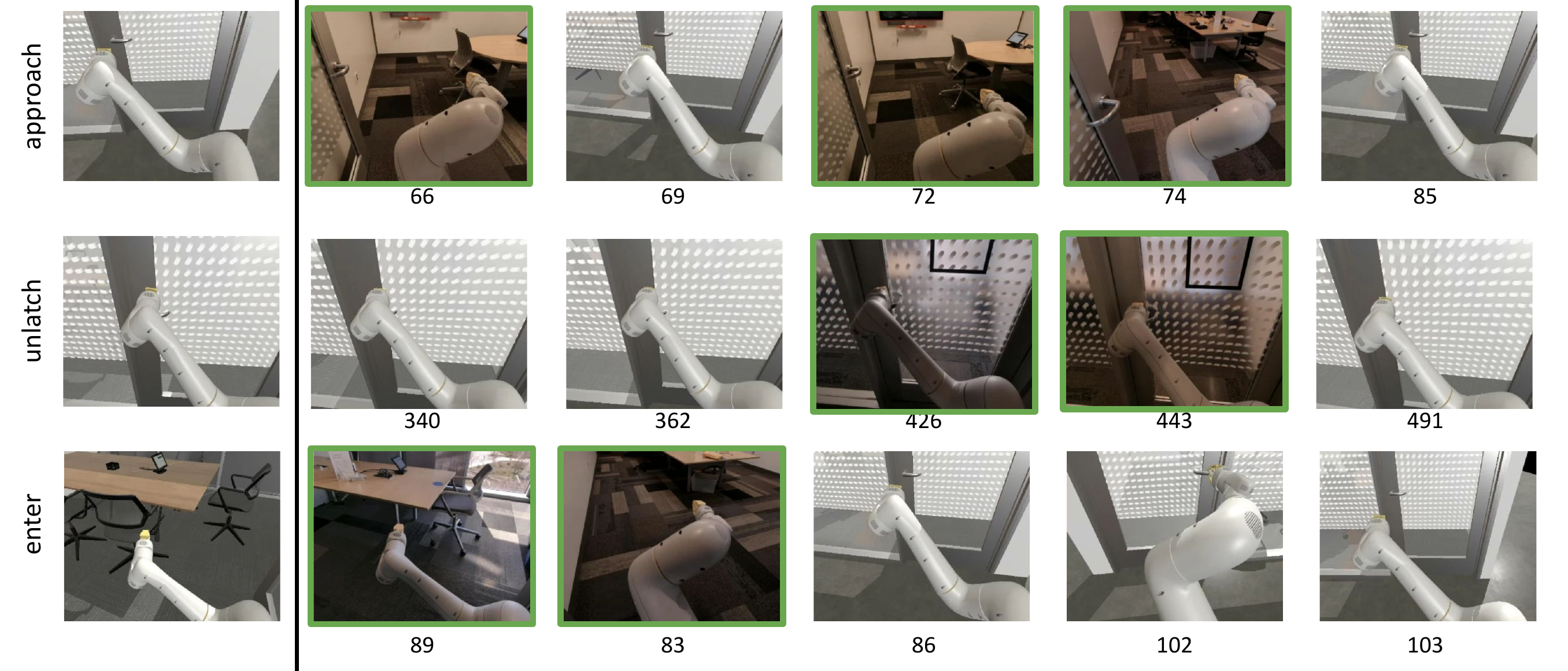}  
  \caption{Simulation anchor images. First row: first phase of door opening moving towards door; second row: manipulating the door handle to unlatch; third row: navigating into the room. Interestingly, the closest images in the first and last rows both show sim/real images in either the approaching phase or navigating phase within the room, likely due to the similar base movement and lack of arm motion. The second row shows similar door unlatching frames across sim/real domains.}
  \label{fig:kl_div_sub_second}
\end{subfigure}
\caption{Anchor images (real) displayed in the leftmost column, with the nearest 5 images to their right. KL divergences (measured relative to the anchor image) given below each similar image. Green borders indicate real images.}
\label{fig:kl_div}
\end{figure*}

In Figure \ref{fig:kl_div}, we look at the nearest neighbours, measured by KL divergence, within a subset of our training dataset consisting of 1600 mixed sim and real images. We look at three primary phases of the task: 1) approaching the door, 2) unlatching and opening the door, and 3) entering the room. Both (1) and (3) only focus on base motion for navigation, while (2) requires a mixture of base and arm motion for manipulation. As (3) includes the interior of the rooms, images from this phase also have greater visual diversity.

In general, we find that the closest images in the embedding space within a KL divergence of $\sim$400 nats correspond to similar actions, which generally corresponds to phases of the door opening task. Notably, the nearest neighbours of simulated images include real images and vice versa, suggesting that our learned representations do not separate images from the two domains when they correspond to similar actions---thus closing the sim-to-real gap. For the corresponding sim-to-real gap investigations in the depth modality, see Appendix \ref{app:depth_gap}.

The first and last rows of Figure \ref{fig:kl_div_sub_second} show an example where the closest images may correspond to a different phase than the anchor image---images anchored by approaching the door show nearest neighbours that are entering the room and vice versa. However, when inspecting the network action predictions for these instances, we find that the actions correspond to the same commands of moving the base forward and keeping the arm still. 
This aligns well with our expectation from the model: the IB objective aims to discard everything that is not relevant for the auxiliary variable---action prediction in our case. The nearest neighbours showing visually different images but corresponding to the same actions suggests that the embedding is discarding anything that is not predictive of the actions.

\textbf{Q2. Model Explainability.} During training, Eq. \ref{eq:bc_vib} aims to minimize the average rate $\mathcal{L}_{KL}$, or KL divergence between the embedding distribution and the learned marginal, to a budget determined by the hyperparameter $\beta$. In doing so, the model must allocate how much divergence is acceptable for each datapoint, which can vary nonuniformly across the dataset. Inputs that are common or correspond to easily predictable actions require less information to encode (lower rates), while the opposite is true for uncommon or hard-to-compress inputs (higher rates). 

We hypothesize that the VIB rate should be higher during portions of the task that are more challenging, as these states likely require more information to act accurately. To investigate this, we plot the change in rate across trajectories for both modalities. A sample is shown in Figure \ref{fig:trajrate}, with additional trajectories of different conditions in Appendix \ref{app:rate_traj}. In general, we find that the rates for both modalities are highest during the most challenging part of the task---manipulating the door latch. This phase is critical, as to successfully open the door, the agent should: control both the arm and the base (9 degree-of-freedom), know which side of the handle to push down on (depending on swing orientation), and take care not to slip off the handle. On the other hand, the rate is lowest for the most intuitively compressible inputs---where the robot only needs to move the base at the start and end of the task (2 degrees-of-freedom). This suggests that the learned prior and encoder are salient to the challenging parts of the task, making the model more explainable.

Furthermore, as shown in Figure \ref{fig:kl_div}, finding the nearest neighbours between a set of anchor images within a subset of the dataset can be used to provide more insights about the quality of a trained model. For example, wrong groupings of images could indicate the bottleneck was too tight for the model to properly distinguish the different phases of the task; or, retrieving both sim and real images for a given anchor image is a healthy sign that the model has learned a domain-agnostic representation.

\textbf{Q3. VIB Rate-based Sensor Fusion.} The previous two experiments suggest that we are capable of 1) learning a representation that helps close the sim-to-real gap, and 2) using the VIB rates to better understand and debug the trained model post hoc. Here, we test our hypotheses that both the reduced sim-to-real gap and VIB rate-based sensor fusion help performance on the real-world task of latched door opening. As observed in other sensor fusion works and our own baselines, combining modalities can be more detrimental than using either modality individually. 

In Table \ref{tab:success_compare}, we compare our method against three baseline models from \cite{khansari2022practical}: two trained on each modality individually and one fusion model that concatenates RGB and depth embeddings together before feeding into the action prediction MLP, a common multi-sensor fusion approach \cite{park2017rdfnet, calandra2018more}. Each baseline has the same architecture as ours, minus the stochastic embedding. We evaluate three fusion methods with VIB: 1) embedding Concatenation Fusion (CF), 2) VIB rate-based action fusion with linear normalization (blended actions), and 3) VIB rate-based action fusion with softmax normalization (modality switching). We find that the domain agnostic representations induced by the bottleneck help improve performance for all three VIB methods above the baselines. In particular, the best VIB methods outperform the best baseline method (RGB only) by 16\%, and the best baseline fusion method by 21\%.  

Interestingly, we find that the RGBD + VIB (CF) method performs equally well as VIB (Softmax Fusion), even though RGBD (CF) was the lowest performing baseline in the real evaluations. However, we also find that the CF method has much higher variance in simulated evaluation performance, suggesting that relying on concatenated embeddings can increase optimization difficulty. See Appendix \ref{app:sim_eval} for corresponding plots. We note that the softmax-based fusion method performs just as well, while additionally having a more interpretable sensor fusion policy by construction. Examining the rates per modality and using the rates to enforce which modality is in control gives assurances as to which sensor the policy is relying on at each point in time, allowing for easier model debugging. Notably, the black box CF policy cannot provide such insights.

Between the rate-based fusion methods, we find that softmax normalization performs slightly better than linear normalization. We hypothesize that this is due to the discrete action selection, whereas the linear method blends the actions using linear interpolation. This may be problematic if each modality has learned to predict different actions for the same state, and the blended action is worse than either action individually.

\begin{table}[t]
  \centering
  {\fontsize{8.6pt}{8.6pt} \selectfont
  \begin{tabular}{@{}l|c|cc@{}}
    \toprule
    Method & Total & Seen & Unseen \\
    \midrule
    RGB  & 80\% $\pm$ 2.3     & 75\% $\pm$ 3.2 & 87\% $\pm$ 3.1\\
    Depth              & 77\% $\pm$ 2.5 & 79\% $\pm$ 3.1 & 75\% $\pm$ 4.2\\
    RGBD (CF)             & 75\% $\pm$ 2.4 & 79\% $\pm$ 3.0 & 69\% $\pm$ 4.3\\
    \hline
    RGBD + VIB (CF)  & 96\% $\pm$ 1.1            & 94\% $\pm$ 1.8 & 98\% $\pm$ 1.3 \\
    RGBD   + VIB (Linear)          & 93\% $\pm$ 1.5 & 93\% $\pm$ 1.9 & 93\% $\pm$ 2.3\\
    RGBD   + VIB (Softmax)           & 96\% $\pm$ 1.1 & 98\% $\pm$ 1.0 & 94\% $\pm$ 2.2\\
    \bottomrule
  \end{tabular}
  }
  \vspace{-3mm}
  \caption{Real-world door opening success rates (\%) $\pm$ standard deviation, based on 300 trials (180 on seen doors, 120 on unseen doors). We compare against baseline models that use each modality individually and a concatenation fusion (CF) model that fuses RGB and depth by concatenating the embeddings together before passing to a shared MLP action decoder. For the VIB models, we compare a CF variant that concatenates bottlenecked embeddings and two rate-based fusion variants using the action fusion schemes described in Section \ref{sec:fusion}. We find that both softmax fusion and CF perform best, with the former having the additional benefit of an explicitly understandable fusion scheme.}
  \label{tab:success_compare}
  \vspace{3mm}
\end{table}

To investigate whether the rate-based sensor fusion is correctly choosing the `better' modality (i.e. that the better performing modality individually has lower rates), we ablate the softmax-based fusion model into its RGB and depth only components in Table \ref{tab:modality_ablation}. Using the same checkpoints, we enforce that the action taken is always from one of the modalities. Interestingly, the depth-only branch performs slightly better than the fused model, while the RGB-only branch performs much worse. Our fused model's performance is close to the depth-only result, suggesting that the rate-based fusion is able to pick out the stronger performing modality and avoid relying on the lower performing modality. We note that the RGB performance here is lower than the baselines due to the checkpoints being originally selected for best softmax-fusion performance in simulation, not best RGB-only performance. 

\begin{table}[t]
  \centering
  {\fontsize{8.6pt}{8.6pt} \selectfont
  \begin{tabular}{@{}l|c|cc@{}}
    \toprule
    Method & Total & Seen & Unseen \\
    \midrule
    RGBD  + VIB Softmax  & 96\% $\pm$ 1.1 & 98\% $\pm$ 1.0 & 94\% $\pm$ 2.2 \\ \hline
    RGB   + VIB          & 74\% $\pm$ 2.5 & 79\% $\pm$ 3.0 & 67\% $\pm$ 4.3 \\
    Depth + VIB          & 98\% $\pm$ .06 & 99\% $\pm$ .07 & 97\% $\pm$ 1.6 \\
    \bottomrule
  \end{tabular}
  }
  \vspace{-3mm}
  \caption{Sensor modality ablation for the softmax-normalized VIB rate-based fusion model. We decompose the contributions of each modality to the Fusion model by using the same model checkpoints, but forcing the model to place all weight on either RGB or Depth. We find that the fused model's performance is closer to the better performing depth-only model, suggesting that the rate-based fusion is able to select the stronger performing modality.}
  \label{tab:modality_ablation}
  \vspace{-3mm}
\end{table}

%% file: conclusions.tex
\section{Conclusions and Future Work}

Motivated by the challenges of closing the multi-sensor sim-to-real gap in end-to-end learning for robotics, we make use of the regularizing capabilities of the VIB to learn a domain agnostic representation. Compared to other common regularization approaches for sim-to-real (eg. domain randomization), the VIB requires no simulation engineering, instead requiring tuning the hyperparameter $\beta$ to find a desirable bottleneck capacity. To combine multiple sensor modalities, our insight is to leverage the VIB's inherent uncertainty estimation for uncertainty-based sensor fusion.  Studying the mobile manipulation task of latched door opening in a real office, we highlight some explainability characteristics of our approach, finding that the rates are salient and that the learned embeddings are sim and real domain agnostic. We evaluate and discuss the tradeoffs between different methods of sensor fusion, significantly improving task performance and successfully deploying the VIB on a real multi-sensor robotic system. 

\textbf{Limitations.} In this paper we assume policy uncertainty correlates with the VIB rates, since the former is a harder metric to measure. 
Under the VIB objective, the encoder is given an average rate budget to allocate across all of the inputs seen. Generally, we should expect common inputs will be economically represented by low rates.  High rate inputs are inputs that the encoder has difficulty compressing.  While it seems reasonable to, and our results indicate that it is appropriate to treat these as examples the policy will perform worse on, this does rely on the encoder and marginal pair being well-aligned.
One possible way to mitigate this issue is to use the tools illustrated in Figures \ref{fig:trajrate} and \ref{fig:kl_div} to examine the quality of the model more directly.

\textbf{Broader Impact.}
By learning an explicit (albeit high-dimensional) density model of data, we have the tools to better understand what a policy has learned (e.g. whether domain-agnostic images from similar portions of the task are similarly encoded under the stochastic embedding). Using the VIB rate as a measure of sensor uncertainty allows the policy to enforce which modality is in control, leading to a more interpretable understanding of which inputs the policy uses to make its decisions. Our hope is that such a capability improves the safety and reliability of robotic systems trained end-to-end with machine learning. Although door opening is a benign application of robotics in and of itself, a potential outcome of this capability is that it unlocks a myriad of indoor robotics applications for which navigating autonomously between rooms was previously very challenging, such as security and patrol inside buildings.

\textbf{Future work.}
In this work, we find that the regularization provided by a bottleneck objective helps to reduce the sim-to-real gap for robotics applications. Other commonly used regularization approaches include using contrastive losses (to learn invariant representations) and/or data augmentations. For future work, it would be interesting to thoroughly investigate and understand the interplay between these types of regularization---whether they provide complementary or redundant regularization capabilities, with what tradeoffs, and how their utilities may vary across learning domains and applications.

%% file: appendix.tex
\appendix
\onecolumn

\section{Detailed Experiment Results}\label{app:exp_results}

\begin{table}[h!]
{\fontsize{8pt}{9pt} \selectfont
\begin{tabular}{l|r|rr|rr|rr|rr|rr|rr}
             & \multicolumn{1}{l|}{Overall} & \multicolumn{2}{c|}{TL1}                                & \multicolumn{2}{c|}{TL2}                                  & \multicolumn{2}{c|}{TR1}                                & \multicolumn{2}{c|}{TR2}                                  & \multicolumn{2}{c|}{TL3}                                     & \multicolumn{2}{c}{TR3}                                     \\ \hline
Robot        & \multicolumn{1}{l|}{}        & \multicolumn{1}{c}{A}       & \multicolumn{1}{c|}{B}    & \multicolumn{1}{c}{A}    & \multicolumn{1}{c|}{B}         & \multicolumn{1}{c}{A}       & \multicolumn{1}{c|}{B}    & \multicolumn{1}{c}{A}    & \multicolumn{1}{c|}{B}         & \multicolumn{1}{c}{A}         & \multicolumn{1}{c|}{B}       & \multicolumn{1}{c}{A}         & \multicolumn{1}{c}{B}       \\ \hline
Time of Day  & \multicolumn{1}{l|}{}        & \multicolumn{1}{c}{\Circle } & \multicolumn{1}{c|}{\LEFTcircle} & \multicolumn{1}{c}{\LEFTcircle} & \multicolumn{1}{c|}{\CIRCLE} & \multicolumn{1}{c}{\Circle } & \multicolumn{1}{c|}{\LEFTcircle} & \multicolumn{1}{c}{\LEFTcircle} & \multicolumn{1}{c|}{\CIRCLE} & \multicolumn{1}{c}{\CIRCLE} & \multicolumn{1}{c|}{\Circle } & \multicolumn{1}{c}{\CIRCLE} & \multicolumn{1}{c}{\Circle } \\ \hline
Lighting     & \multicolumn{1}{l|}{}        & \multicolumn{1}{c}{On}      & \multicolumn{1}{c|}{Off}  & \multicolumn{1}{c}{Off}  & \multicolumn{1}{c|}{On}        & \multicolumn{1}{c}{On}      & \multicolumn{1}{c|}{Off}  & \multicolumn{1}{c}{Off}  & \multicolumn{1}{c|}{On}        & \multicolumn{1}{c}{On}        & \multicolumn{1}{c|}{Off}     & \multicolumn{1}{c}{Off}       & \multicolumn{1}{c}{On}      \\ \hline
RGB    & 75\%                        & 100\%                       & 73\%                     & 47\%                     & 73\%                          & 93\%                        & 7\%                      & 87\%                     & 100\%                         & 73\%                          & 80\%                        & 80\%                          & 87\%                        \\
Depth  & 79\%                        & 80\%                        & 100\%                    & 60\%                     & 100\%                         & 87\%                        & 53\%                     & 100\%                    & 53\%                          & 27\%                          & 100\%                       & 100\%                         & 87\%                        \\
RGBD (CF)  & 79\%                        & 53\%                        & 93\%                     & 40\%                     & 100\%                         & 100\%                       & 40\%                     & 100\%                    & 100\%                         & 33\%                          & 93\%                        & 100\%                         & 100\%       \\ \hline  

RGBD + VIB (CF)   & 94\%                        & 87\%                        & 100\%                     &73\%                     & 100\%                         & 87\%                       & 100\%                     & 100\%                    & 100\%                         & 100\%                      & 100\%                             & 87\%                        & 93\%      \\
RGBD + VIB (Linear)   & 93\%                        & 80\%                        & 100\%                     &87\%                     & 93\%                         & 93\%                       & 100\%                     & 100\%                    & 100\%                         & 80\%                      & 100\%                             & 80\%                        & 100\%  \\
RGBD + VIB (Softmax)   & 98\%                        & 100\%                        & 100\%                     &100\%                     & 93\%                         & 100\%                       & 100\%                     & 93\%                    & 100\%                         & 100\%                      & 100\%                             & 87\%                        & 100\%  
\end{tabular}
}
\caption{Full results for training doors, broken down by robot, time of day [\Circle \;Morning, \LEFTcircle \;Noon, \CIRCLE \;Afternoon], and lighting conditions.}
\label{tab:full_train}
\end{table}

\begin{table}[h!]
\centering
\begin{tabular}{l|r|rr|rr|rr|rr}
             & \multicolumn{1}{l|}{Overall} & \multicolumn{2}{c}{ER1}                                     & \multicolumn{2}{c}{EL1}                                     & \multicolumn{2}{c}{EL2}                                & \multicolumn{2}{c}{ER2}                                  \\ \hline
Robot        & \multicolumn{1}{l|}{}        & \multicolumn{1}{c}{A}       & \multicolumn{1}{c|}{B}         & \multicolumn{1}{c}{A}       & \multicolumn{1}{c|}{B}         & \multicolumn{1}{c}{A}       & \multicolumn{1}{c|}{B}    & \multicolumn{1}{c}{A}    & \multicolumn{1}{c}{B}         \\ \hline
Time of Day  & \multicolumn{1}{l|}{}        & \multicolumn{1}{c}{\Circle} & \multicolumn{1}{c|}{\CIRCLE} & \multicolumn{1}{c}{\Circle} & \multicolumn{1}{c|}{\CIRCLE} & \multicolumn{1}{c}{\Circle} & \multicolumn{1}{c|}{ \LEFTcircle} & \multicolumn{1}{c}{ \LEFTcircle} & \multicolumn{1}{c}{\CIRCLE} \\ \hline
Lighting     & \multicolumn{1}{l|}{}        & \multicolumn{1}{c}{On}      & \multicolumn{1}{c|}{Off}       & \multicolumn{1}{c}{Off}     & \multicolumn{1}{c|}{On}        & \multicolumn{1}{c}{On}      & \multicolumn{1}{c|}{Off}  & \multicolumn{1}{c}{Off}  & \multicolumn{1}{c}{On}        \\ \hline
RGB     & 87\%                        & 100\%                       & 100\%                         & 47\%                        & 80\%                          & 80\%                        & 93\%                     & 100\%                    & 93\%                          \\
Depth  & 75\%                        & 100\%                       & 47\%                          & 60\%                        & 60\%                          & 40\%                        & 93\%                     & 100\%                    & 100\%                          \\
RGBD  (CF)   & 69\%                        & 93\%                        & 47\%                          & 53\%                        & 40\%                          & 87\%                        & 67\%                     & 93\%                     & 73\%  \\ \hline
RGBD + VIB (CF)   & 98\%                        & 100\%                        & 93\%                          & 100\%                        & 100\%                          & 93\%                        & 100\%                     & 100\%                     & 100\%  \\
RGBD + VIB (Linear)   & 93\%                        & 100\%                        & 100\%                          & 80\%                        & 100\%                          & 93\%                        & 80\%                     & 100\%                     & 93\%  \\
RGBD + VIB (Softmax)   & 93\%                        & 100\%                        & 93\%                          & 87\%                        & 100\%                          & 100\%                        & 100\%                     & 80\%                     & 80\% 
\end{tabular}
\caption{Full results for evaluation doors, broken down by robot, time of day [\Circle \;Morning, \LEFTcircle \;Noon, \CIRCLE \;Afternoon], and lighting conditions.}
\label{tab:full_eval}
\end{table}

\newpage
\section{Depth Sim-to-Real Gap}\label{app:depth_gap}
\begin{figure*}[h!]
\begin{subfigure}{\linewidth}
  \centering
  % include first image
  \includegraphics[width=\linewidth]{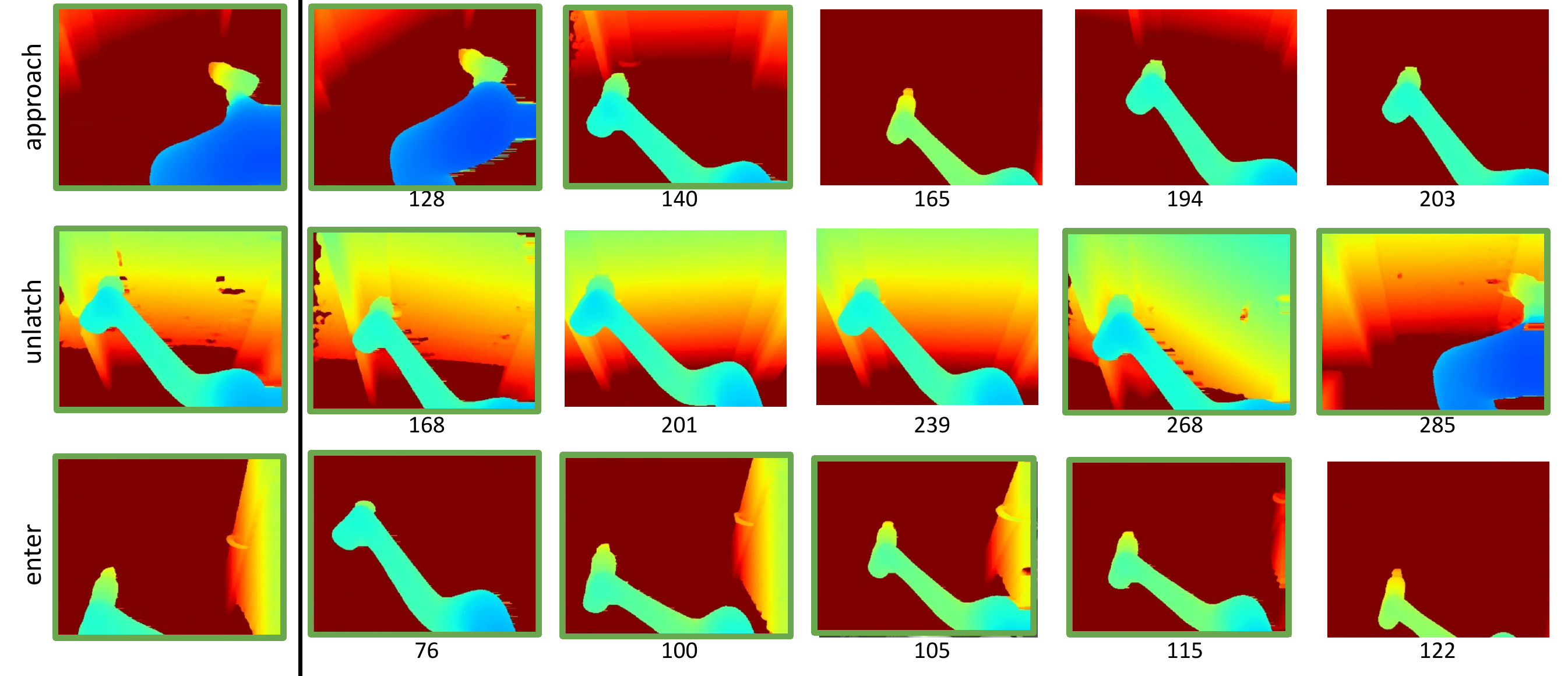}  
  \caption{First row: first phase of door opening moving towards door, second row: manipulating door handle to unlatch, third row: navigating into the room. Real anchor images. The closest images in the first row are all during the approaching phase, varying in arm orientation and showing both sim/real domains. The second row shows the same door unlatching configuration, both sim/real domains. The last row shows different room interiors, both sim/real domains. Green outline indicates real depth images.}
  \label{fig:sub-first-1}
\end{subfigure}
\begin{subfigure}{\linewidth}
  \centering
  % include second image
  \includegraphics[width=\linewidth]{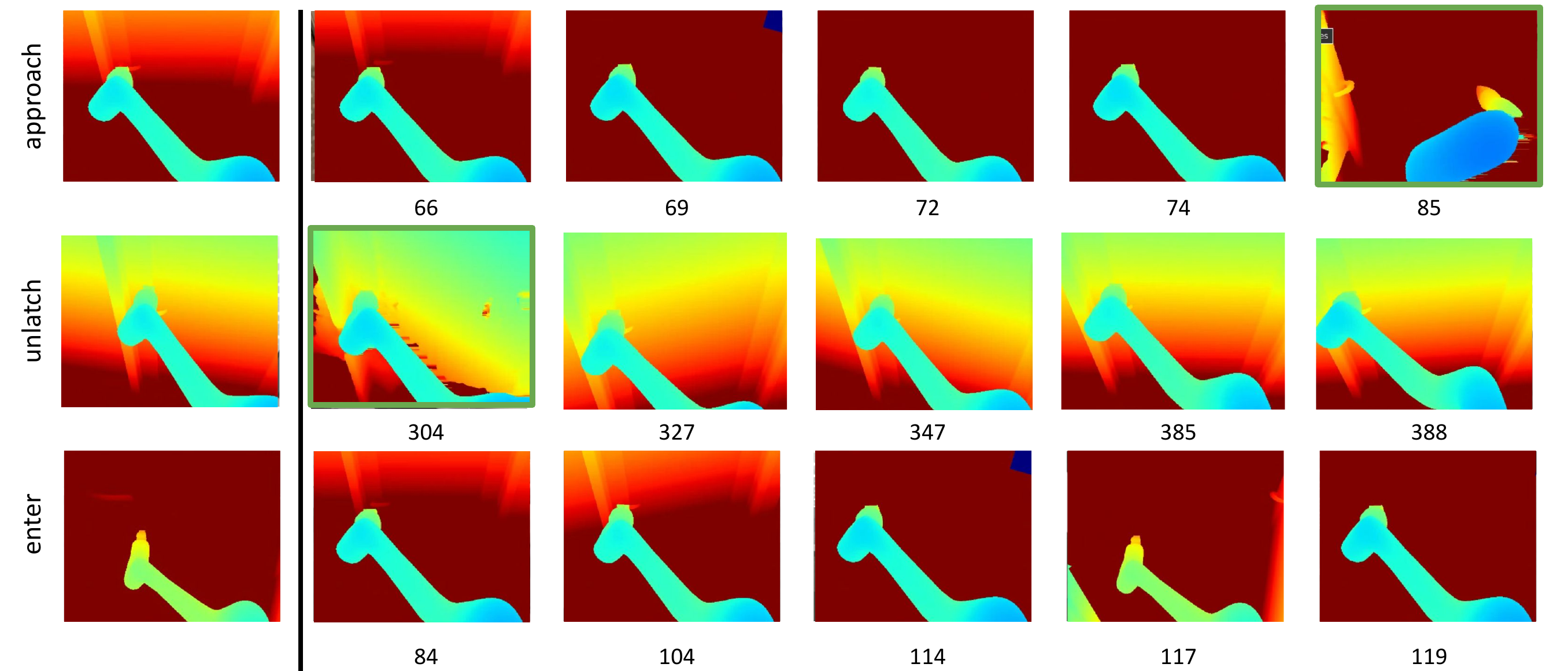}  
  \caption{First row: first phase of door opening moving towards door, second row: manipulating door handle to unlatch, third row: navigating into the room. Simulated anchor images. Interestingly, the closest images in the first and last rows both show sim/real images in either the approaching phase or navigating phase within the room, likely due to the similar base movement. The second row shows similar door unlatching frames across sim/real domains. Green outline indicates real depth images.}
  \label{fig:sub-second-1}
\end{subfigure}
\caption{ Anchor image (real) on the left, with the nearest 5 images on the right. KL divergences (measured relative to the anchor image) given below each similar image. }
\label{fig:kl_div_app}
\end{figure*}
\newpage

\section{Additional Rate Trajectories}\label{app:rate_traj}
\begin{figure*}[ht!]
\begin{subfigure}{.5\linewidth}
  \centering
  % include first image
  \includegraphics[width=\linewidth]{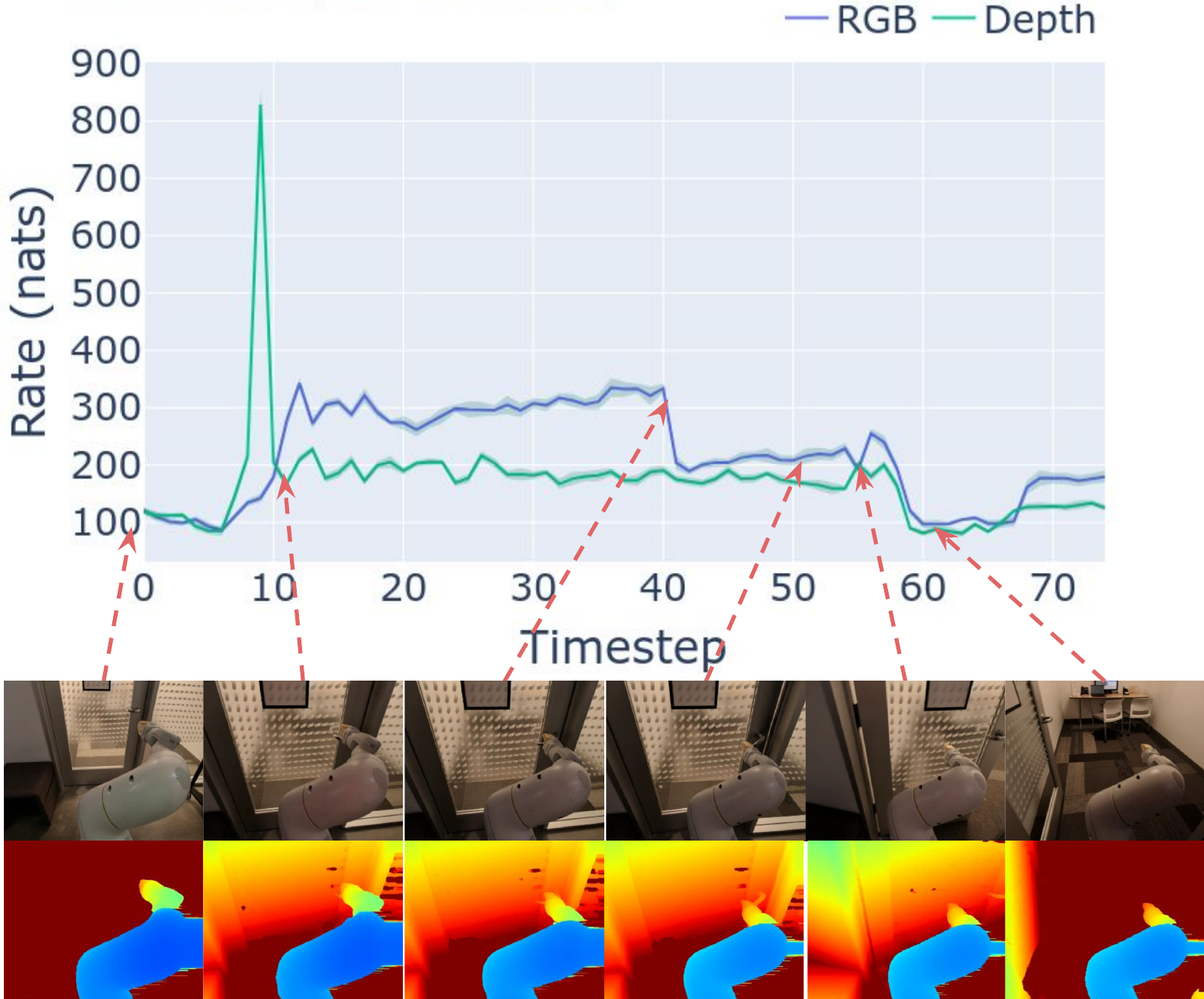}  
  \caption{Sample success episode with a right-swing open door. }
  \label{fig:sub-first-2}
\end{subfigure}
\begin{subfigure}{.5\linewidth}
  \centering
  % include second image
  \includegraphics[width=\linewidth]{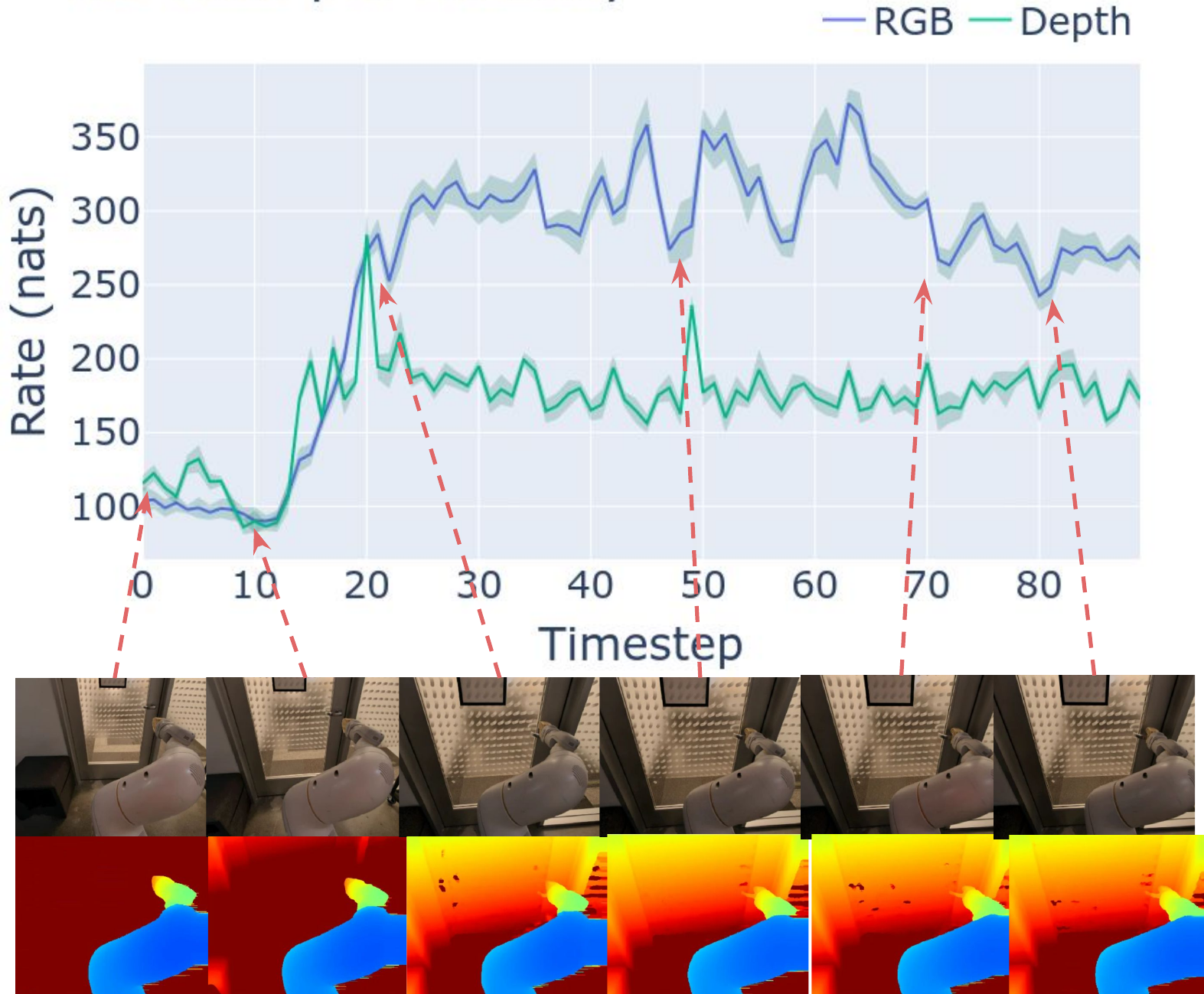}  
  \caption{Sample failure episode, where we see the rate remains elevated as the gripper tries to open the door but fails due to exerting force on the wrong side of the latch.}
  \label{fig:sub-second-2}
\end{subfigure}
\begin{subfigure}{.5\linewidth}
  \centering
  % include first image
  \includegraphics[width=\linewidth]{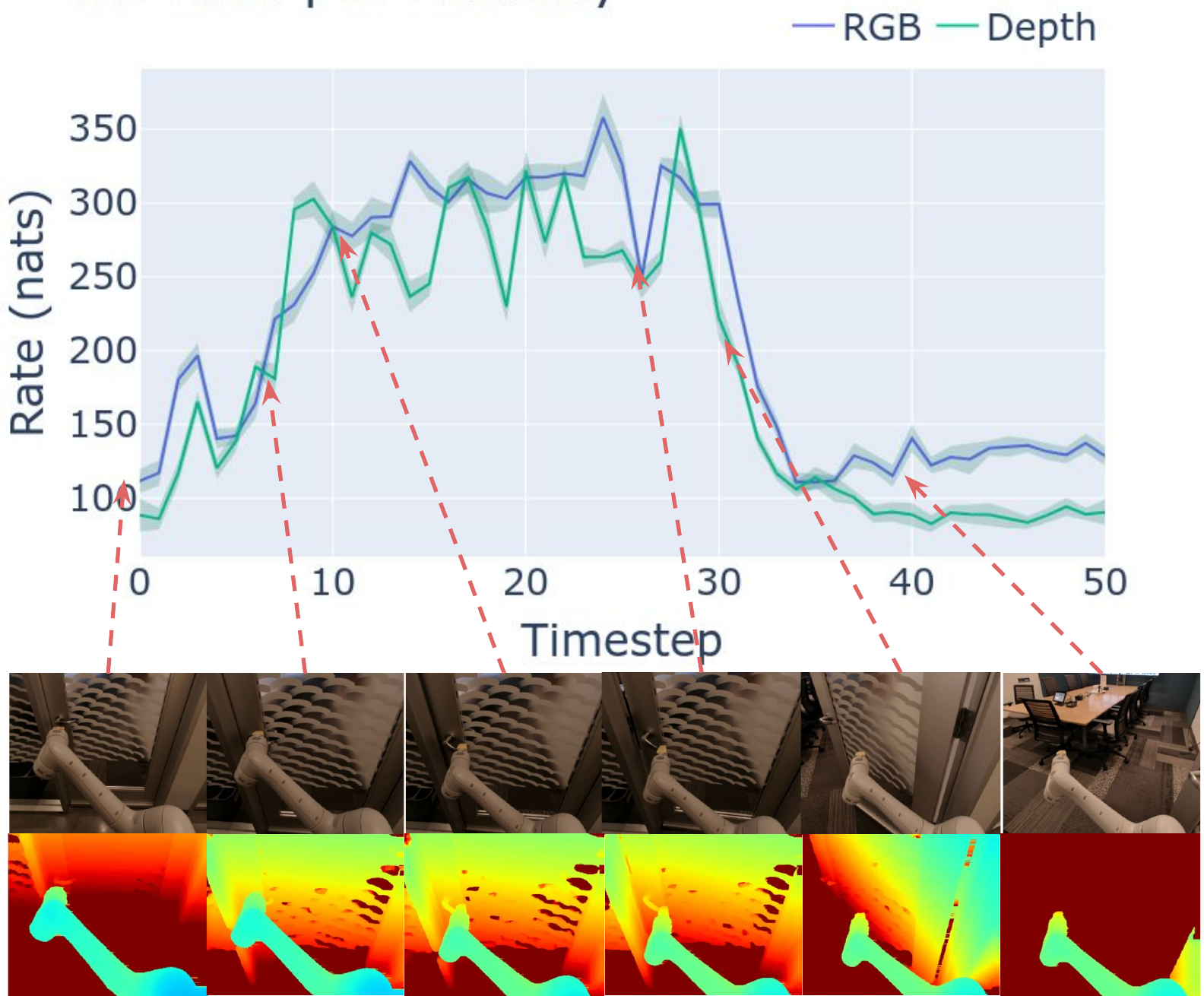}  
  \caption{Sample success episode with a differently patterned door, not seen during training.}
  \label{fig:sub-first-3}
\end{subfigure}
\begin{subfigure}{.5\linewidth}
  \centering
  % include second image
  \includegraphics[width=\linewidth]{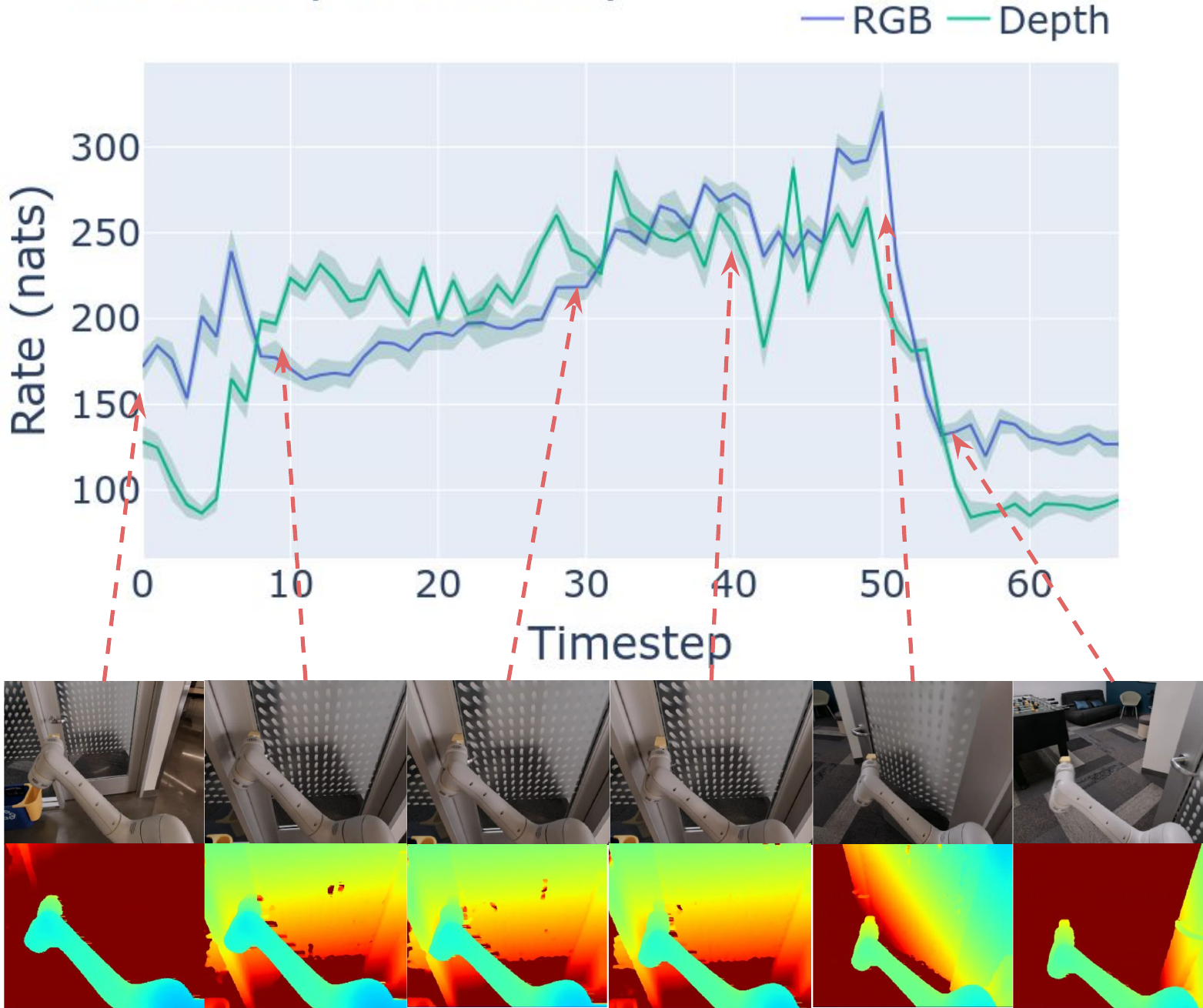}  
  \caption{Sample success episode with a differently coloured interior.}
  \label{fig:sub-second-3}
\end{subfigure}
\caption{Rates per modality across additional sample trajectories. Plots show VIB Rate (nats) over time, with corresponding RGB (blue) and depth (green) images labelled with red arrows. In general, the rates are highest during the critical door unlatching phase.}
\label{fig:rate_traj_app_1}
\end{figure*}

\begin{figure*}[ht!]
\begin{subfigure}{.5\linewidth}
  \centering
  % include first image
  \includegraphics[width=\linewidth]{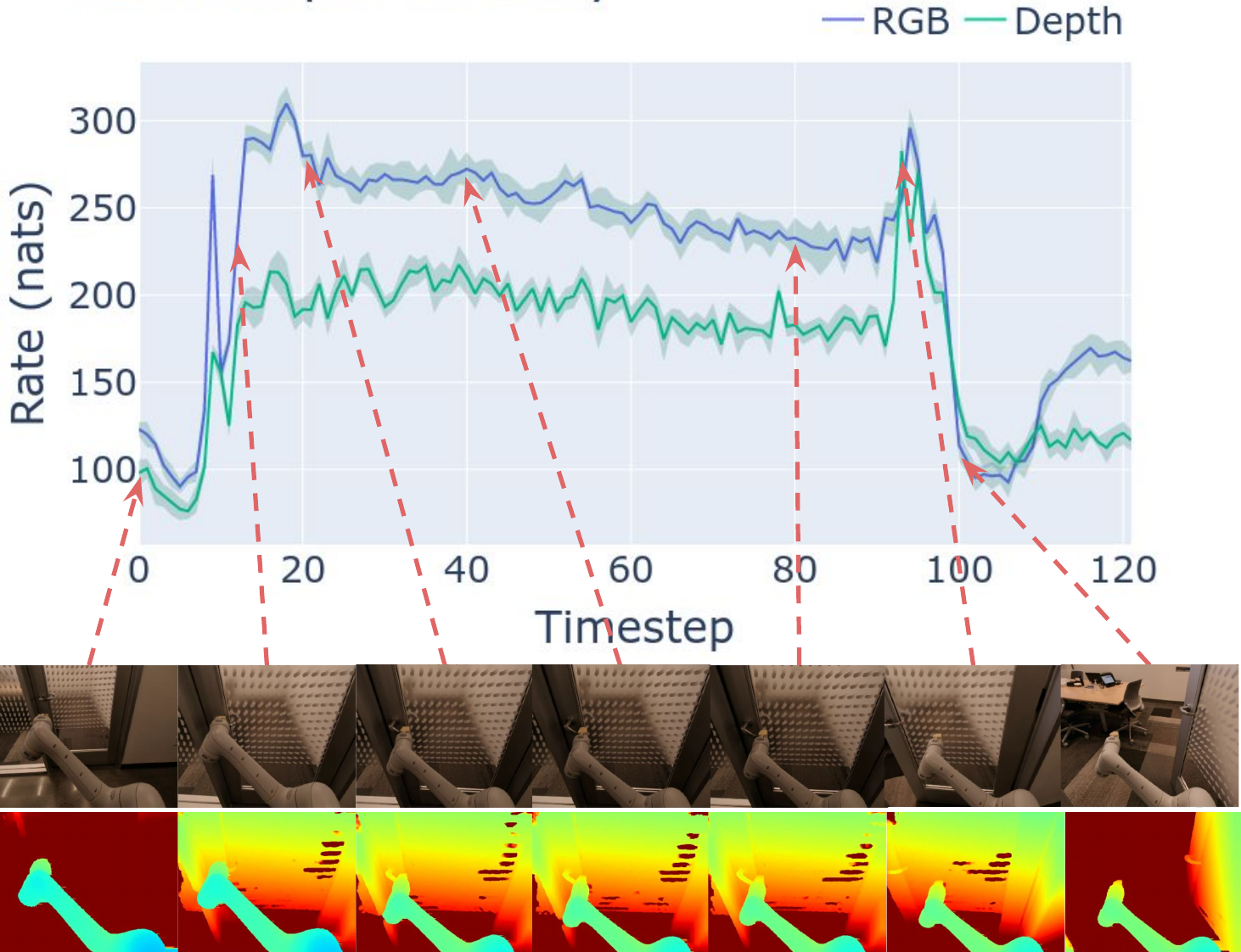}
  \caption{. }
  \label{fig:sub-first-4}
\end{subfigure}
\begin{subfigure}{.5\linewidth}
  \centering
  % include second image
  \includegraphics[width=\linewidth]{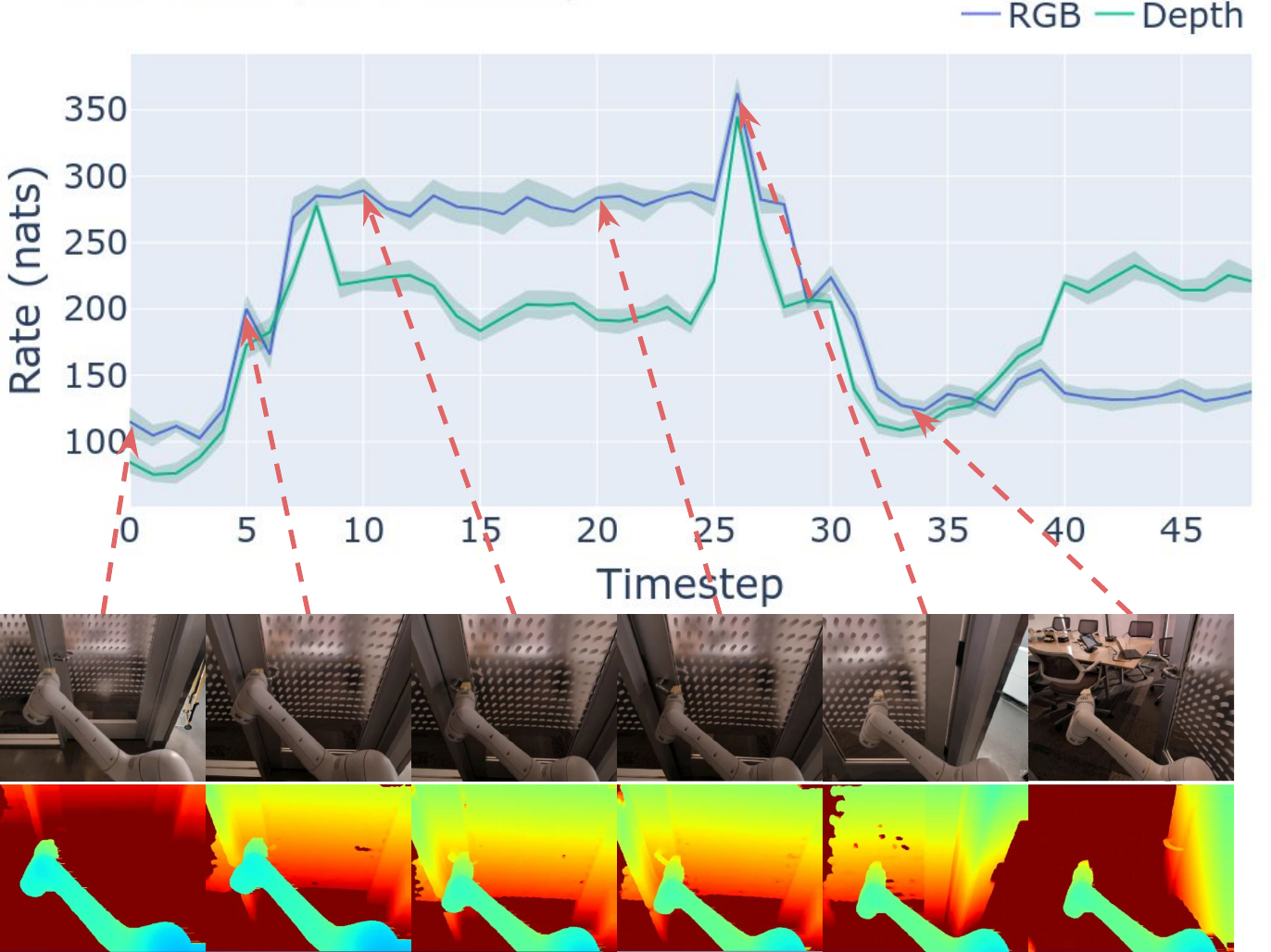} 
  \caption{}
  \label{fig:sub-second-4}
\end{subfigure}
\begin{subfigure}{.5\linewidth}
  \centering
  % include first image
  \includegraphics[width=\linewidth]{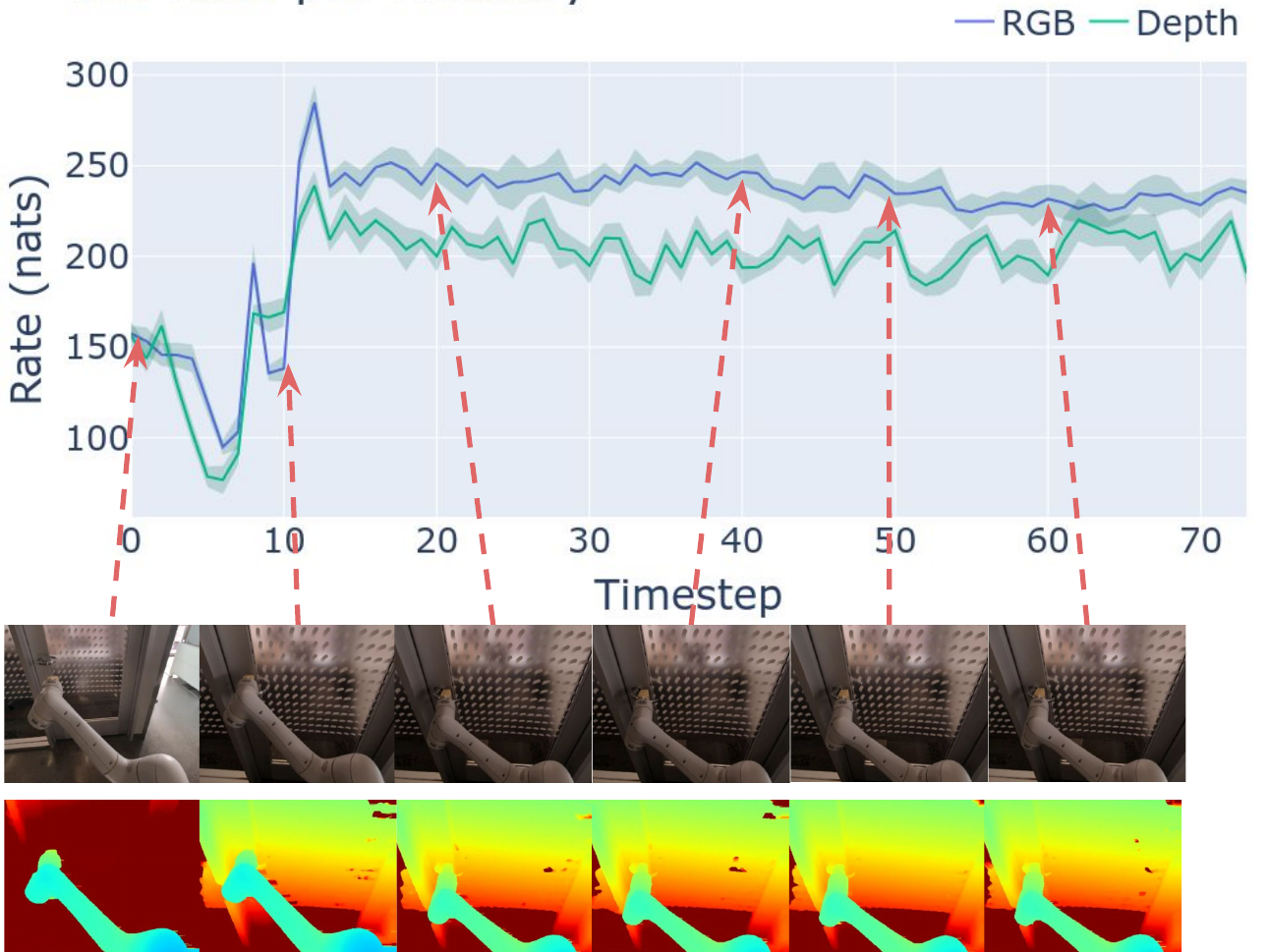}
  \caption{}
  \label{fig:sub-first-5}
\end{subfigure}
\begin{subfigure}{.5\linewidth}
  \centering
  % include second image
  \includegraphics[width=\linewidth]{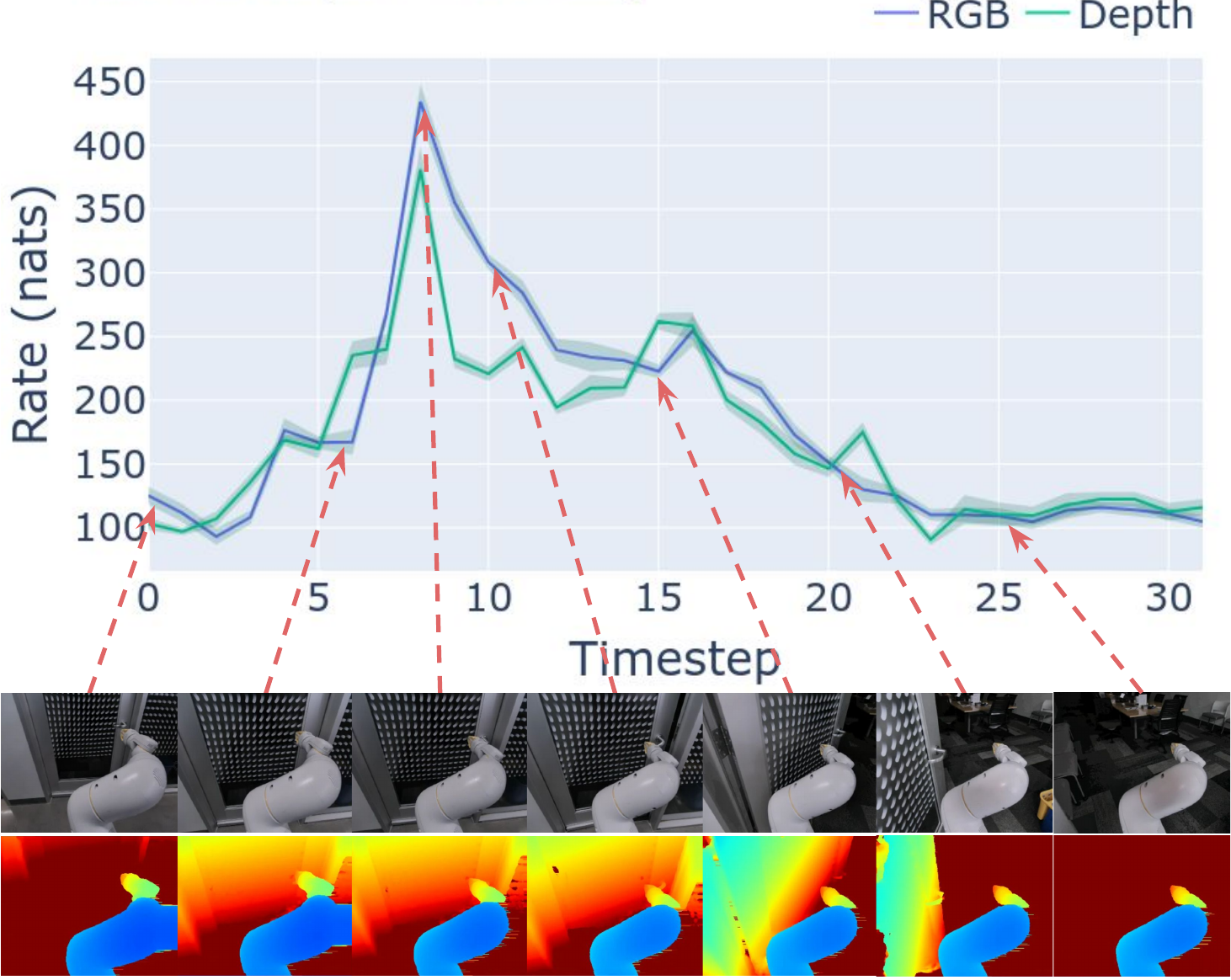}
  \caption{}
  \label{fig:sub-second-5}
\end{subfigure}
\caption{Rates per modality across additional sample trajectories. Plots show VIB Rate (nats) over time, with corresponding RGB (blue) and depth (green) images labelled with red arrows. In general, the rates are highest during the critical door unlatching phase.}
\label{fig:rate_traj_app_2}
\end{figure*}

\clearpage
\section{Simulated Evaluations}\label{app:sim_eval}

\begin{figure*}[ht!]
\begin{subfigure}{\linewidth}
  \centering
  % include second image
  \includegraphics[width=.75\linewidth]{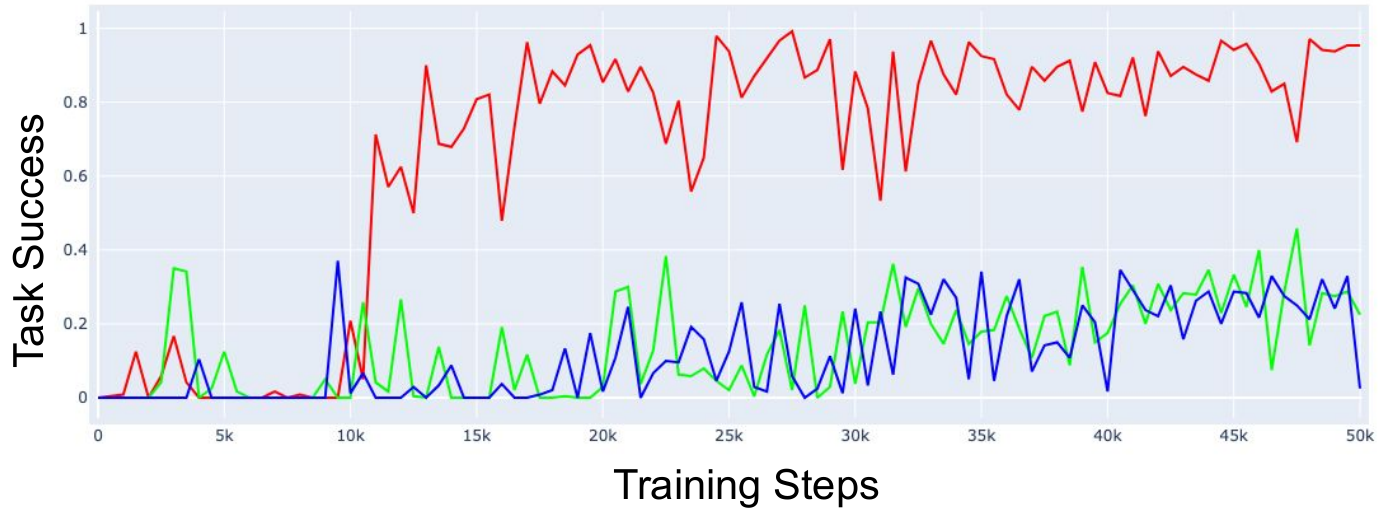}  
  \caption{Simulated evaluation performance for 3 seeds trained with fusion via concatenating RGB and depth embeddings.}
  \label{fig:sub-second-6}
\end{subfigure}
\begin{subfigure}{\linewidth}
  \centering
  % include first image
  \includegraphics[width=.75\linewidth]{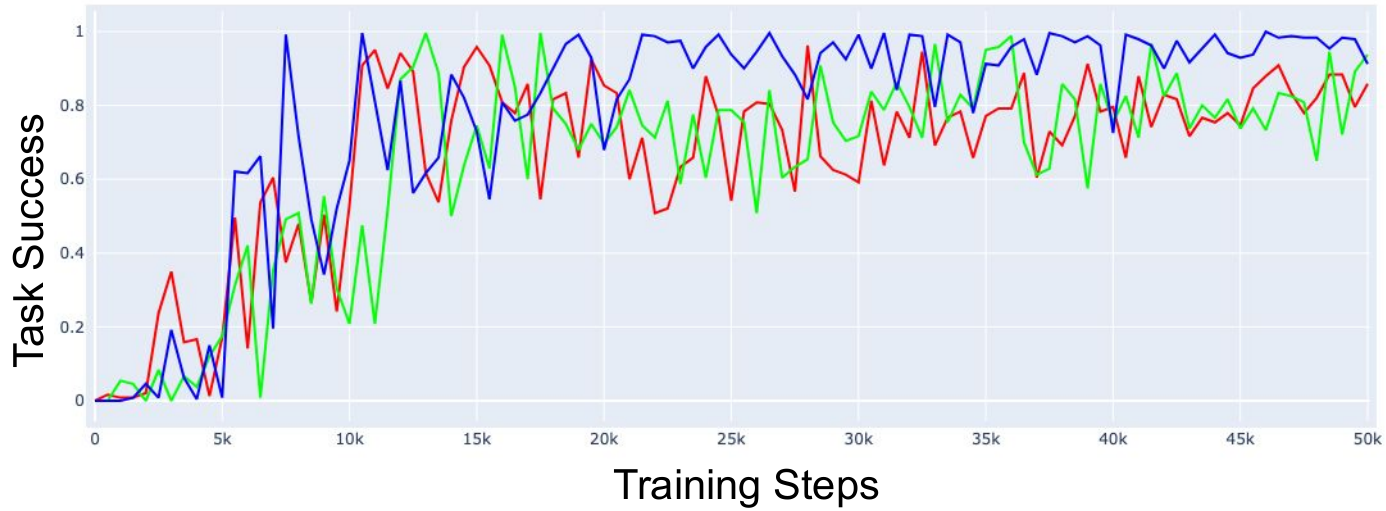}  
  \caption{Simulated evaluation performance for 3 seeds trained with fusion via linearly normalized fusion.}
  \label{fig:sub-first-6}
\end{subfigure}
\begin{subfigure}{\linewidth}
  \centering
  % include second image
  \includegraphics[width=.75\linewidth]{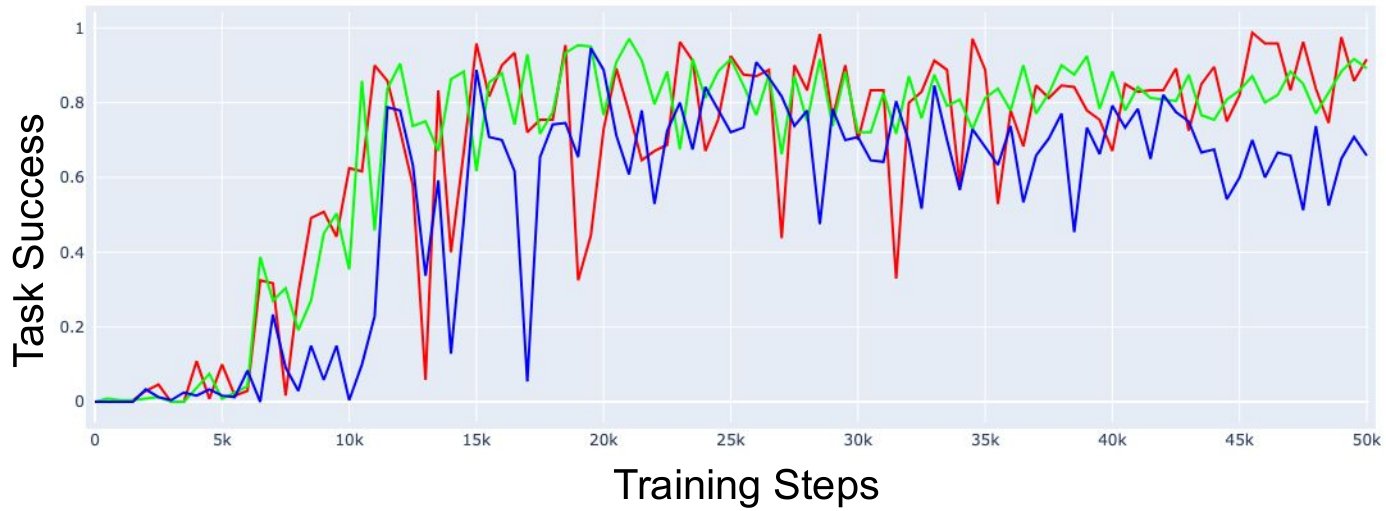}  
  \caption{Simulated evaluation performance for 3 seeds trained with fusion via softmax normalized fusion.}
  \label{fig:sub-third}
\end{subfigure}
\caption{Simulated evaluation success rates on 6 seen simulated rooms throughout training. For real evaluations we choose the highest performing checkpoints from sim. Notably, the concatenated fusion model is harder to train, with greater variance between seeds.}
\label{fig:sim_success}
\end{figure*}